%% file: ijcai22.tex
\documentclass{article}
\usepackage{ijcai22}

\usepackage{times}
\usepackage{soul}
\usepackage{url}
\usepackage[hidelinks]{hyperref}
\usepackage[utf8]{inputenc}
\usepackage[small]{caption}
\usepackage{graphicx}
\usepackage{amsmath}
\usepackage{amsthm}
\usepackage{booktabs}
\usepackage{algorithm}
\usepackage{algorithmic}
\usepackage{bm}
\usepackage{amssymb}
\usepackage{amsfonts}
\usepackage{xspace}
\usepackage{booktabs, multirow} 
\usepackage{soul}
\usepackage[table]{xcolor} 
\usepackage{changepage,threeparttable} 
\usepackage{color}
\usepackage{comment}

\newcommand{\iwasawa}[1]{{\color{black}{#1}}}
\newcommand{\kojima}[1]{{\color{black}{#1}}}

\newcommand{\lm}{ViT\xspace}

\title{Robustifying Vision Transformer without Retraining from Scratch \\ by Test-Time Class-Conditional Feature Alignment}

\author{
\kojima{
Takeshi Kojima\footnote{\kojima{Contact Author}}\and
Yutaka Matsuo\And
Yusuke Iwasawa\\
}
\affiliations
\kojima{
The University of Tokyo, Japan\\
}
\emails
\kojima{
\{t.kojima,matsuo,iwasawa\}@weblab.t.u-tokyo.ac.jp
}
}

\begin{document}

\maketitle

\begin{abstract}

Vision Transformer (ViT) is becoming more popular in image processing. Specifically, we investigate the effectiveness of test-time adaptation (TTA) on ViT, a technique that has emerged to correct its prediction during test-time by itself. First, we benchmark various test-time adaptation approaches on ViT-B16 and ViT-L16. It is shown that the TTA is effective on ViT and the prior-convention (sensibly selecting modulation parameters) is not necessary when using proper loss function. Based on the observation, we propose a new test-time adaptation method called class-conditional feature alignment (CFA), which minimizes both the class-conditional distribution differences and the whole distribution differences of the hidden representation between the source and target in an online manner. Experiments of image classification tasks on common corruption (CIFAR-10-C, CIFAR-100-C, and ImageNet-C) and domain adaptation (digits datasets and ImageNet-Sketch) show that CFA stably outperforms the existing baselines on various datasets. We also verify that CFA is model agnostic by experimenting on ResNet, MLP-Mixer, and several ViT variants (ViT-AugReg, DeiT, and BeiT). Using BeiT backbone, CFA achieves 19.8\% top-1 error rate on ImageNet-C, outperforming the existing test-time adaptation baseline 44.0\%. This is a state-of-the-art result among TTA methods that do not need to alter training phase.
\footnote{Code is available at \url{https://github.com/kojima-takeshi188/CFA}.}
\end{abstract}

\section{Introduction}
\kojima{Inspired by} the success in natural language processing, Transformer \cite{vaswani2017attention} is becoming more and more popular in various image processing tasks, \kojima{including image recognition \cite{dosovitskiy2020image,touvron2021training}, object detection \cite{carion2020end}, and video processing \cite{zhou2018end,zeng2020learning}.}
Notably, \cite{dosovitskiy2020image} proposed Vision Transformer (ViT), which adapts Transformer architecture to image classification tasks and shows that it achieves comparable or superior performance to that of the conventional convolutional neural networks (CNNs). 
Follow-up research also shows that ViT is more robust to the common corruptions and perturbations than convolution-based models (e.g., ResNet) \cite{paul2021vision,morrison2021exploring}, which is an important property for safety-critical applications. 

This study seeks to answer the following question: \textit{can we improve the robustness of \lm without retraining it from scratch}? 
Most prior works focused on how to robustify the models during training. 
For example, \cite{hendrycks2019augmix,hendrycks2020many} demonstrated that several data augmentation improves robustness of convolutional neural networks (CNN). 
Similarly, \cite{chen2021vision} shows that a sharpness-aware optimizer improves the robustness of ViT. 
Unfortunately, such approaches require retraining the models from scratch, which entails a massive computational burden and training time for large models (such as ViT). 
Moreover, sometimes dataset for pre-training is not publicly available, which makes it impossible to retrain the models. 

This study investigates the effectiveness of test-time adaptation (TTA) to robustify ViT.  
TTA is a recently emerged approach for improving the robustness of models without retraining them from scratch and accessing to training dataset \cite{schneider2020improving,nado2021evaluating,wang2020tent}. 
Instead, it corrects the model's prediction for test data by modulating its parameters during test time. 
For example, \cite{wang2020tent} proposed Tent, which modulates the parameters of batch normalization (BN) by minimizing prediction entropy. It was shown that Tent can significantly improve the robustness of ResNet. 
TTA has two major advantages over usual training-time techniques. 
First, it does not alter the training phase and thus does not need to repeat the computationally heavy training phase. 
Second, it does not require accessing to the source data during adaptation, which is impossible in the case of large pre-trained models. 

Conceptually speaking, TTA can be applied to arbitrary network architectures. 
However, naively modulating model parameters during test-time may cause a catastrophic failure as discussed in \cite{wang2020tent}. 
To avoid the issue, prior works often limited the modulation parameters, which resulted in architecture constraints.
For example, \cite{schneider2020improving,wang2020tent} modulated statistics and/or affine transformation in batch normalization (BN) layer, but the BN-based method cannot be applied to some modern models such as ViT since they do not have BN.
 

This study contributes to addressing this research question by following two means. 
First, we benchmark various test-time adaptation methods on ViT using several robustness benchmark tasks (CIFAR10-C, CIFAR100-C, ImageNet-C, and several domain adaptation tasks). We also design modulation parameters potentially suitable for ViT-based architectures.
The tested methods include entropy minimization  \cite{wang2020tent}, pseudo-classifier  \cite{iwasawa2021testtime}, pseudo-label  \cite{lee2013pseudo}, diversity regularization  \cite{liang2020we}, and feature alignments  \cite{liu2021ttt}. 
Regarding modulation parameters, we sweep over the following four candidates: layer normalization \cite{ba2016layer}, CLS token, feature extractor \cite{liang2020we}, and entire parameters of a model. 
The results indicate that the prior convention in test-time adaptation (i.e., limiting modulation parameters) \textit{is not necessary} when using proper loss function, while it is necessary for pure entropy minimization based approach. 
This observation is important for applying TTA to arbitrary network architectures. 

We then propose a new loss function: test-time class-conditional feature alignment (CFA).
Our approach can be categorized into feature alignment approach as with \cite{liu2021ttt}, which minimize the gap of the statistics between training and test domain. 
\iwasawa{It is worth noting that the feature alignment approach for test-time adaptation (e.g., our approach and \cite{liu2021ttt}) assumes that one can access to the statistics on the source dataset during the test phase but does not need to access to the source dataset itself and to repeat the computationally heavy training} 
\footnote{\iwasawa{Test-time adaptation generally assumes that the model would be distributed without source data due to bandwidth, privacy, or profit reasons \cite{wang2020tent}. We argue that the statistics of source data would be distributed even in such a situation since it could drastically compress data size and eliminate sensitive information. In fact, some layers often used in typical neural networks contain statistics of source data (e.g., batch normalization)}}.  
Therefore, this approach can be used without source dataset during adaptation. 
We show that such complementary information about the source data distribution can stabilize the training without selecting the modulation parameters.
In addition, we extend the feature alignment approach \cite{liu2021ttt} by the following two means.
First, CFA aligns class-conditional statistics \kojima{as well as} the statistics of overall distribution.
Second, we calculate the statistics after properly normalizing the hidden representations.
Despite the simplicity, these techniques significantly boost the performance of test-time adaptation.


In summary, our main contributions are as follows.
\begin{itemize}
\item This is the first study that verifies the effectiveness of test-time adaptation methods on \lm. By benchmarking several test-time adaptation approaches under \kojima{common} corruptions and domain adaptation tasks, 
we have validated that the robustness of \lm model is improved during test time without retraining the model from scratch. 
\item 
We introduce a new test-time adaptation method (CFA). Throughout the experiment, CFA achieves better results than existing baselines on multiple datasets. In addition, CFA is robust to hyperparameter tuning, which is important in practically setting up test-time adaptation.
\item We show that CFA consistently improves the robustness on a wide variety of backbone networks during test time. 
In particular, we achieve the state-of-the-art results of test-time adaptation on ImageNet-C with a 19.8 \% top-1 error rate when using BeiT-L16 as a backbone network.
\end{itemize}

\section{Related Work}
\subsection{Vision Transformer (ViT)}
\label{sec_transformer_architecture}
Transformer \cite{vaswani2017attention}, first proposed in natural language processing (NLP) field,  also achieves great performance in image processing as Vision Transformer (ViT) \cite{dosovitskiy2020image}. ViT divides input image data into small patches and translates them to embedding vectors, otherwise known as a "token". Extra learnable class embedding (CLS token) is added to the sequence of the tokens before feeding them into Transformer Encoder. 
Transformer Encoder mainly consists of multilayered global self-attention blocks and MLP blocks. The blocks include layer normalization \cite{ba2016layer} as one function. 
An MLP head is added to top layer of the CLS token as a classifier.

Since its invention, ViT has rapidly become popular in the field of computer vision. Many applications and extensions have been proposed thus far.
For example, \cite{touvron2021training},\cite{bao2021beit}, and \cite{steiner2021train} showed that the performance of ViT is respectively improved by distilation (DeiT), self-supervised learning (BeiT), and data augumentation (ViT-AugReg) during pre-training phase. 
More recently, \cite{tolstikhin2021mlp} proposed MLP-Mixer, which was proven to be quite competitive with ViT by replacing self-attention blocks with MLP layers.

This study is interested in how to robustify ViT to the common perturbations. 
Recent experimental research has verified that ViT inherently has robustness without any adaptation or any additional data augmentation. 
Several studies empirically show that ViT is inherently more robust than CNNs \cite{paul2021vision,morrison2021exploring,naseer2021intriguing,Yamada_2022_CVPR} by using some benchmark datasets.
Several studies have shown that the robustness of ViT can be improved by changing the training strategy, such as using a larger data set for the pre-training phase \cite{paul2021vision,bhojanapalli2021understanding} or a sharpness-aware optimizer for the training phase \cite{chen2021vision}. 
However, retraining such a massively pre-trained model from scratch is not desirable considering the computational burden. 
The larger the data and model, the higher it costs for retraining.
At the same time, however, the model size also matters for robustness, i.e., larger models tend to be more robust by themselves 
(See Appendix E for the detail). 
This observation motivated us to investigate a lightweight and model-agnostic way to improve the robustness of the models. 



\subsection{Test-Time Adaptation (TTA)}
\label{sec_test_time_adaptation}
This study investigates the effectiveness of the test-time adaptation approaches for Vision Transformer and its variants. 
Unlike most existing works that focus on training phase to improve the robustness, test-time adaptation focuses on test-time. 
In other words, test-time adaptation does not alter the training phase; therefore, we do not need to repetitively run computationally heavy training to improve robustness. 

The algorithm of existing test-time adaptation can be summarized by following two aspects: (1) adaptation function $f_{adapt}$ and (2) modulation parameters $\bm{\psi}$. 
Literally, $f_{adapt}$ is a function that determines how to modulate the model parameter during the test time. 
More formally, $f_{adapt}$ receives a batch of unlabeled images $\bm{X}_{test}$, which is available online at test-time, and updates the target parameter $\bm{\psi}$ using the data. 
A naive instance of $f_{adapt}$ might use stochastic gradient decent (SGD) by designing a loss function that can effectively incorporate $\bm{X}_{test}$ to correct its prediction. 
For example, Tent \cite{wang2020tent}, which is a pioneering method for test-time adaptation, minimizes prediction entropy using SGD, based on the assumption that a more confident prediction (i.e. low prediction entropy) leads to a more accurate prediction. 
One can also use different loss functions (such as pseudo-label (PL) \cite{lee2013pseudo}, diversity regularization (SHOT-IM) \cite{liang2020we}, and feature alignments (TFA) \cite{liu2021ttt}), or design optimization-free procedures to update the model (e.g., T-BN \cite{schneider2020improving,nado2021evaluating} and T3A \cite{iwasawa2021testtime}). 

The second aspect is the selection of modulation parameters $\bm{\psi}$. 
As discussed in \cite{wang2020tent}, updating the entire model parameters $\bm{\theta}$ is often ineffective in test-time optimization because $\bm{\theta}$ is usually the only information of the source data in the setup, and updating all parameters without restriction results in catastrophic failure. (See Table \ref{table_modulation_study} for the experiment result). 
Consequently, prior works also proposed to sensibly select modulation parameters along with the adaptation method $f_{adapt}$. 
For example, \cite{schneider2020improving,nado2021evaluating} proposed to re-estimate the statistics of batch normalization \cite{ioffe2015batch} during the test time while fixing the other parameters.  %
Similarly, Tent \cite{wang2020tent} modulated only a set of affine transformation parameters of the BN layer. 
This causes two problems when applying test-time adaptation to ViT. 
First, ViT has significantly larger parameters compared to ResNet which is the standard test bed of prior studies. 
Consequently, the effectiveness of TTA on such a huge model has not been fully investigated. 
Second, ViT and its variants do not have BN, so they cannot directly take advantage of the common good strategy.
In other words, there is a lack of knowledge regarding which parameter should be updated to effectively robustify ViT.

In this study, we avoid the difficulty of sensibly selecting the modulation parameters by incorporating the feature-alignment approach. 
More specifically, we explicitly minimize the difference between some statistics of source distribution and target (test) distribution, rather than simply modulating model parameters only given data from target distribution. 
In other words, we leverage the source statistics as auxiliary information regarding the source distribution to prevent adaptation from causing the aforementioned catastrophic failure. 
Note that our method does not rely on the co-existence of source and target data and does not violate the setting of test-time adaptation.

Similar to our work, \cite{liu2021ttt} recently proposed test-time feature alignment (TFA), which aligns the hidden representation between source and target data by minimizing the distance of the mean and covariance matrix. 
Our method is different from TFA in the following two aspects. 
First, we propose to align class-conditional statistics \kojima{as well as} the statistics of overall distribution. 
Second, we propose to calculate the statistics after properly normalizing the hidden representations. 
In \S 4.4, our experiment results demonstrate that these techniques stably improve the performance of various tasks based on various backbone networks.

\section{Methods}
\subsection{Modulation Parameters}
\label{sec_tent}
As discussed in \S \ref{sec_test_time_adaptation}, the choice of modulation parameters is regarded as important in test-time adaptation but prior BN-based modulation is not applicable to Vision Transformer. 
To find good candidates for $\bm{\psi}$ in ViT, we sweep over the following four candidates: layer normalization \cite{ba2016layer}, CLS token, feature extractor parameters, and all parameters.

A Layer normalization (LN) re-estimates the mean and standard deviation of input across the dimensions of the input, followed by the affine transformation for each dimension. 
We update the affine transformation parameters in LN for adaptation. 
A CLS token is a parameterized vector and proven to be efficient for fine-tuning large models for downstream tasks in NLP \cite{lester2021power}. \kojima{A feature extractor is defined as any module in a model except for its classifier. This term is borrowed from \cite{liang2020we}. In the case of ViT, its feature extractor consists of Transformer Encoder, patch embeddings, and positional embeddings.}
Updating feature extractor parameters is a basic unsupervised domain adaptation setting, while \cite{wang2020tent} claimed that it was ineffective in test-time adaptation setup (see \S 2.2). 

It is worth noting that these choices of modulation parameters are applicable to many modern architectures, including ViT, DeiT, MLP-Mixer, and BeiT. 
This property is important in practice because a better backbone network usually provides significant performance gains. 
We also empirically show the effectiveness of TTA on such various architectures. 

\subsection{Class-Conditional Feature Alignment}

Regarding the adaptation function, this study proposes a new loss function, called class-conditional feature alignment (CFA). 
Similar to the most prior works, our method uses stochastic gradient decent to adapt the model during test-time. 
Unlike the prior methods such as Tent, PL, and T3A that modulate the parameters using the data available at test-time only, our method aligns the statistics of features between source and target. 
In other words, we leverage the source statistics as an auxiliary information regarding the source distribution to prevent the model from suffering a catastrophic failure. 


Assume that a model consists of two components; a linear classifier $g_{\omega}$ as last layer, and a feature extractor $f_{\phi}$ before the classifier. 
A set of source training samples is denoted as $X^s=\{x^s_i\}^{N_s}_{i=1}$.
While prior works often calculate the statistics of feature output by $f_{\phi}$, the feature is not always normalized. For example, ViT uses GELU as an activation function
and LN with elementwise affine transformation before classifier, which is not bounded. 
We found that this causes unstable behavior especially when matching higher order moments of distributions. 
Thus, before calculating the statistics, we normalize (bound the minimum and maximum value of) the hidden representation for each sample $f_{\phi}(x^s_i)$ as follows.
\begin{align}
h(x^s_i) =& Tanh \left( LN^{\dagger} \left( f_{\phi}(x^s_i) \right) \right), 
\label{equ_h_std_s}
\end{align}
where $LN^{\dagger}$ is defined as layer normalization 
\textit{without affine transformation}.
Despite the simplicity, we empirically find that \kojima{not only matching higher order moment of overall distribution is stabilized, but also} the performance of class-conditional feature alignment is boosted (See Table \ref{table_ablation_study} for the detail). 
The feature normalization might have a positive effect on class-conditional distribution matching by highlighting the distribution property of each class.

\begin{algorithm}[t]
\caption{Online Adaptation using CFA}
\label{alg:algorithm}
\textbf{Input}: 
Fine-tuned DNN model with parameters $\theta$, Partial parameters to be updated during adaptation $\psi \subset \theta$,
Target test dataset $X^{t}$,
$m$-th ordered batch data $ X^{t,m} \subset X^{t}$,
Statistics of Eq.(\ref{equ_statistics_s2}) (\ref{equ_statistics_s3}) (\ref{equ_statistics_s1}) calculated from source training dataset.
\\
\textbf{Output}:
\begin{algorithmic}[1] 
\FOR{$m=1$ to M}
\STATE Predict labels $\hat{Y}^{t,m}$ for $ X^{t,m}$
\STATE Calculate statistics Eq.(\ref{equ_statistics_t2}) (\ref{equ_statistics_t3}) (\ref{equ_statistics_t1}) for $ X^{t,m}$
\STATE Update $\psi$ using Eq.(\ref{equ_total_loss})
\ENDFOR
\RETURN $(\hat{Y}^{t,1},.., \hat{Y}^{t,M})$
\end{algorithmic}
\end{algorithm}

After the normalization, the mean and higher order central moments of overall distribution on source data are calculated and stored in memory as fixed values.
\begin{align}
\mathbb{\mu}^s =& \frac{1}{|X^{s}|}\sum_{x^{s}_i \in X^{s}} {h(x^{s}_i)}, \label{equ_statistics_s2}\\ 
\mathbb{M}^s_k =& \frac{1}{|X^{s}|}\sum_{x^{s}_i \in X^{s}}(h(x^s_i)-\mathbb{\mu}^s)^{k}, 
\quad (k = 2,...,K)
\label{equ_statistics_s3}
\end{align}
where $K$ denotes the maximum number of moments. 
Class-conditional mean of the normalized hidden representations is also calculated and stored in memory as fixed value as follows
\begin{align}
\mathbb{\mu}^s_c =& \frac{1}{|X^{s}_{c}|}\sum_{x^{s}_i \in X^{s}_c} {h(x^{s}_i)}, 
\hspace{46pt} (c = 1,...,C)
\label{equ_statistics_s1}
\end{align}
where $C$ denotes the number of classes. $X^s_c \subset X^s$ contains all the source samples whose ground-truth labels are c. 
Note that these statistics are calculated before adaptation, i.e., we do not need to access to source data itself in test phase.

CFA uses these statistics to adapt the model during test phase. 
\kojima{Assume that a sequence of test data
drawn from target distribution arrives at our model one after another.}
Test dataset is denoted as $X^t=\{x^t_i\}^{N_t}_{i=1}$, and
a set of test data in $m$-th batch is denoted as
$X^{t,m} \subset X^{t}, (m = 1,...,M)$. 
For each batch, 
hidden representations of test data are normalized and their statistics are calculated in the same way as source. 
\begin{align}
h(x^t_i) = Tanh \left( LN^{\dagger} \left( f_{\phi}(x^t_i) \right) \right), 
\label{equ_h_std_t}
\end{align}
\begin{align}
\mathbb{\mu}^{t,m} =& \frac{1}{|X^{t,m}|}\sum_{x^{t}_i \in X^{t,m}} {h(x^{t}_i)}, \label{equ_statistics_t2}\\ 
\mathbb{M}^{t,m}_k =& \frac{1}{|X^{t,m}|}\sum_{x^{t}_i \in X^{t,m}}(h(x^t_i)-\mathbb{\mu}^{t,m})^{k}, 
\: (k = 2...K)
\label{equ_statistics_t3}\\
\mathbb{\mu}^{t,m}_c =& \frac{1}{|X^{t,m}_{c}|}\sum_{x^{t}_i \in X^{t,m}_c} {h(x^{t}_i)}, 
\hspace{40pt} (c = 1,...,C)
\label{equ_statistics_t1}
\end{align}
where $X^{t,m}_c$, which is a subset of $X^{t,m}$, includes all samples in the current batch annotated as class c by pseudo-labeling $argmax_c \: g_{\omega}(f_{\phi}(x^t_i))$. In this study, the overall distribution distance is defined by the central moment distance (CMD) \cite{CMD}
(see Appendix G for details):
\begin{align}
\mathcal{L}_F =& \frac{1}{2} || \mathbb{\mu}^s - \mathbb{\mu}^{t,m} ||_2 + \frac{1}{2^k} \sum_{k=2}^K || \mathbb{M}^s_k - \mathbb{M}^{t,m}_k ||_2. 
\label{equ_f_loss}
\end{align}
As for class-conditional distribution matching, \kojima{following prior studies in UDA setting (\cite{xie2018learning,deng2019cluster}),} we use class-conditional centroid alignment.
\begin{align}
\mathcal{L}_C =& \frac{1}{2|C^{\prime}|} \sum_{c \in C^{\prime}} || \mathbb{\mu}^s_c - \mathbb{\mu}^{t,m}_c ||_2, 
\label{equ_c_loss}
\end{align}
where $C^{\prime}$ denotes a set of the pseudo labelled classes belonging to the current target minibatch samples. 
The first-order moment (centroid) is sufficient for class-conditional feature alignment when class size is larger than batch size.
Parameters $\psi$ of the model are updated by the gradient of the following loss function based on the target batch data at hand.
\begin{align}
\mathcal{L} =& \mathcal{L}_F + \lambda \mathcal{L}_C,
\label{equ_total_loss}
\end{align}
where $\lambda$ is a balancing hyperparameter. Following \cite{wang2020tent}, for efficient computation, we use the scheme that the parameter update follows the prediction for the current batch. Therefore, the update only affects the next batch. The adaptation procedure is summarized in Algorithm \ref{alg:algorithm}.

\begin{table}[t]\centering
\begin{tabular}{lrrrrr}\toprule
ViT-B16&Tent &PL &SHOT-IM &CFA \\
\midrule
LN &\textbf{50.6±0.5} &\textbf{55.7±1.4} &45.7±0.0 &43.9±0.0 \\
CLS &59.4±0.0 &60.6±0.0 &59.9±0.0 &58.2±0.0 \\
Feature &56.2±2.2 &60.8±2.1 &\textbf{43.9±0.0} &\textbf{41.8±0.0} \\
ALL &59.1±1.0 &61.4±2.2 &44.0±0.0 &\textbf{41.8±0.0} \\
\midrule
ViT-L16 &Tent &PL &SHOT-IM &CFA \\
\midrule
LN &\textbf{42.3±0.0} &\textbf{44.3±0.0} &42.0±0.0 &40.2±0.0 \\
CLS &50.3±0.0 &51.3±0.0 &50.7±0.1 &49.2±0.0 \\
Feature &43.8±0.6 &46.5±0.8 &\textbf{38.4±0.0} &\textbf{36.6±0.0} \\
ALL &44.2±1.1 &46.9±0.7 &\textbf{38.4±0.0} &\textbf{36.6±0.0} \\
\bottomrule
\end{tabular}
\caption{Modulation parameter choice study. The evaluation metric is top-1 error on ImageNet-C averaged over 15 corruption types with severity level of 5. ViT-B16 and ViT-L16 are used as the models. CLS: CLS token, LN: layernorm params, Feature: parameters of feature extractor, ALL: all the parameters of ViT.}
\label{table_modulation_study}
\end{table}

\begin{table*}[t]\centering
\begin{tabular}{lcc|rrrrrr}\toprule
& & Class &C10$\rightarrow$ &C100$\rightarrow$ &ImageNet$\rightarrow$ &SVHN$\rightarrow$ &SVHN$\rightarrow$ &ImageNet$\rightarrow$\\
Method & Type & Cond.&C10-C &C100-C &ImageNet-C &MNIST &MNIST-M &ImageNet-S\\
\midrule
Source & - & &14.6±0.0 &35.1±0.0 &61.9±0.0 &23.2±0.0 &46.2±0.0 &64.1±0.0\\
T3A & gf & &13.7±0.0 &34.0±0.0 &61.2±0.0  &17.4±0.3 &40.9±0.2 &61.7±0.0\\
\midrule
Tent & fm & &10.9±0.2 &27.4±0.5 &50.6±0.5 &15.3±0.2 &53.0±1.7 &68.3±4.3\\
PL & fm & &11.9±0.0 &30.1±0.5 &55.7±1.4 &15.8±0.7 &49.7±1.9 &62.2±1.2\\
SHOT-IM & fm &  &8.9±0.0 &25.6±0.0 &45.7±0.0 &\textbf{13.7±0.1} &36.6±0.4 &\textbf{56.1±0.1}\\
\midrule
TFA(-) & fa & &8.8±0.0 &32.2±0.2 &57.8±0.1 &16.5±0.1 &39.3±0.4 &65.7±0.2\\
CFA-F (Ours) & fa &  &\textbf{8.7±0.0} &\textbf{25.2±0.0} &46.7±0.0 &16.3±0.0 &39.9±0.1 &57.2±0.1\\
\midrule
CFA-C (Ours) & fa & \checkmark&\textbf{8.5±0.0} &\textbf{25.3±0.1} &\textbf{45.3±0.0} &14.2±0.0 &\textbf{35.8±0.2} &57.6±0.1\\
CFA (Ours) & fa & \checkmark&\textbf{8.4±0.0} &\textbf{24.6±0.1} &\textbf{43.9±0.0} &14.2±0.1 &\textbf{36.3±0.2} &\textbf{56.1±0.0}\\
\bottomrule
\end{tabular}
\caption{Method comparison on each adaptation tasks. The evaluation metric is top-1 error rate. The results of CIFAR-10-C, CIFAR-100-C and ImageNet-C are ones averaged over the 15 corruption types with highest severity level (=5). CFA-F : Overall distribution matching only. CFA-C :  Class-conditional distribution matching only. gf : Gradient free method. fm : Method that controls the output feature representation (without depending on feature alignment) by modulation. fa : Method that utilizes feature alignment between source and target by modulation. Our proposal (CFA-C and CFA) is the only method utilizing the class-conditional feature alignment during test-time adaptation. }
\label{table_method_comparison}
\end{table*}

\section{Experiment}
\label{sec:experiment}

\subsection{Datasets and Task Design}
\paragraph{\kojima{Common} Corruptions.} We validate the robustness against \kojima{common} corruptions on CIFAR-10-C, CIFAR-100-C and ImageNet-C \cite{hendrycks2019robustness} as target datasets. These datasets contain data with 15 types corruptions with five levels of severity, that is, each dataset has 75 distinct corruptions. Most of our experiments use the highest severity(=5) datasets as they can make the difference in performance most noticeable.
As source datasets, CIFAR-10, CIFAR-100 \cite{Krizhevsky_2009_17719} and ImageNet(-2012) \cite{russakovsky2015imagenet} are used, respectively.

\paragraph{Domain Adaptation.} We validate the robustness against style shift on small-sized datasets and large-sized datasets. For small-sized datasets, we evaluate the adaptation from SVHN to MNIST / MNIST-M \cite{svhn,lecun1998gradient,ganin2015unsupervised}. For large-sized datasets, we evaluate the adaptation from ImageNet to ImageNet-Sketch \cite{wang2019learning}. 
See Appendix A for a detail description of each dataset.

\subsection{Implementation Details}
\label{sec_adapt_setting}
Vistion Transformer (ViT-B16) is used as a default model throughout the experiment unless an explicit explanation is provided. Images of all the datasets are resized to $224{\times}224$ 
(see Appendix B for details).
Before adaptation, the model is fine-tuned on each source dataset 
(see Appendix C for details). 
In addition, the central moments statistics of hidden representation based on source data need to be calculated to store them in memory. For this purpose, we use all the training data in the source dataset and set the dropout \cite{srivastava2014dropout} off in the model during the calculation. 

As for default hyperparameters for adaptation on target data, batch size is set as 64, optimizer is set as SGD with a constant learning rate of 0.001, and momentum of 0.9 with gradient clipping \cite{Zhang2020Why} at global norm 1.0 across all the experiments (Gradient clipping has the effect of preventing adaptation by Tent from catastrophic failure in the severe corruption setting. 
See ``Ablation Study'' in \S \ref{sec_quantitative_result}).
As for CFA, the balancing parameter $\lambda$ is set as 1.0, and maximum central moments order K is set as 3. During prediction and parameter update, dropout is set off in models.

As an evaluation metric, top-1 error of classification is used across all the experiments. We run all the experiments three times with different seeds for different data ordering by shuffling. A mean and unbiased standard deviation of the metric are reported. Our implementation is in PyTorch \kojima{\cite{paszke2019pytorch}. We use various backbone networks from timm library \cite{rw2019timm} and torchvision library
(Appendix D)}.
Every experiment is run on cloud A100 x 1GPU instance.

\subsection{Baseline Methods}
We compare CFA with some existing baseline test-time adaptation methods \kojima{that do not need to alter training phase} as described in \S\ref{sec_test_time_adaptation}: \textbf{Tent, PL, TFA(-)}\footnote{\kojima{Original TFA needs to alter training phase (add contrastive learning), while this study focuses on robustifying large-scale models without retraining them from scratch. Therefore,} we have changed some of the settings from the original TFA \kojima{so that the model does not need to alter training phase. The modified version of TFA is denoted as TFA(-) in our experiments.} 
See Appendix F for details.}
, \textbf{T3A and SHOT-IM}. In addition, we report the performance of the model on target datasets without any adaptation as \textbf{Source}. T-BN is excluded from the baseline because some models (ViT variants and MLP-Mixer) do not have a batch normalization layer. For a fair comparison, we use the same hyperparameters across all the methods as described in \S\ref{sec_adapt_setting}. 

\begin{table}[t]\centering
\begin{tabular}{lrrr}\toprule
&ImageNet-C &ImageNet-S \\\midrule
ResNet50 &82.0\scriptsize{±0.0} &75.4\scriptsize{±0.0} \\
\small{+ CFA / SHOT-IM} &\textbf{58.8\scriptsize{±0.0}}/\textbf{58.8\scriptsize{±0.0}} &70.0\scriptsize{±0.2}\normalsize{/}\textbf{69.2\scriptsize{±0.1}} \\
ResNet101 &77.4\scriptsize{±0.0} &72.3\scriptsize{±0.0} \\
\small{+ CFA / SHOT-IM} &\textbf{55.3\scriptsize{±0.1}}/55.7\scriptsize{±0.0} &66.8\scriptsize{±0.0}\normalsize{/}\textbf{66.2\scriptsize{±0.1}} \\
\midrule
ViT-B16 &61.9\scriptsize{±0.0} &64.1\scriptsize{±0.0} \\
\small{+ CFA / SHOT-IM} &\textbf{43.9\scriptsize{±0.0}}/45.7\scriptsize{±0.0} &\textbf{56.0\scriptsize{±0.1}}/56.1\scriptsize{±0.1} \\
ViT-L16 &53.4\scriptsize{±0.0} &59.1\scriptsize{±0.0} \\
\small{+ CFA / SHOT-IM} &\textbf{40.2\scriptsize{±0.0}}/42.0\scriptsize{±0.0} &\textbf{52.6\scriptsize{±0.0}}/53.6\scriptsize{±0.1} \\
\midrule
DeiT-S16 &59.9\scriptsize{±0.0} &66.6\scriptsize{±0.0} \\
\small{+ CFA / SHOT-IM} &\textbf{46.0\scriptsize{±0.0}}/46.1\scriptsize{±0.0} &60.3\scriptsize{±0.1}\normalsize{/}\textbf{59.4\scriptsize{±0.0}} \\
DeiT-B16 &52.9\scriptsize{±0.0} &62.5\scriptsize{±0.0} \\
\small{+ CFA / SHOT-IM} &\textbf{39.9\scriptsize{±0.0}}/\textbf{39.9\scriptsize{±0.0}} &55.9\scriptsize{±0.0}\normalsize{/}\textbf{55.4\scriptsize{±0.0}} \\
\midrule
MLP-Mixer-B16 &73.3\scriptsize{±0.0} &74.3\scriptsize{±0.0} \\
\small{+ CFA / SHOT-IM} &\textbf{52.4\scriptsize{±0.1}}/55.1\scriptsize{±0.1} &\textbf{64.2\scriptsize{±0.1}}/65.9\scriptsize{±0.2} \\
MLP-Mixer-L16 &77.1\scriptsize{±0.0} &79.8\scriptsize{±0.0} \\
\small{+ CFA / SHOT-IM} &\textbf{56.3\scriptsize{±0.0}}/62.4\scriptsize{±0.1} &\textbf{70.8\scriptsize{±0.3}}/72.9\scriptsize{±0.3} \\
\midrule
ViT-B16-AugReg &49.0\scriptsize{±0.0} &57.0\scriptsize{±0.0} \\
\small{+ CFA / SHOT-IM} &\textbf{37.6\scriptsize{±0.0}}/38.4\scriptsize{±0.0} &51.5\scriptsize{±0.1}\normalsize{/}\textbf{51.0\scriptsize{±0.2}} \\
ViT-L16-AugReg &39.1\scriptsize{±0.0} &48.2\scriptsize{±0.0} \\
\small{+ CFA / SHOT-IM} &\textbf{32.1\scriptsize{±0.0}}/33.3\scriptsize{±0.0} &\textbf{45.2\scriptsize{±0.0}}/45.6\scriptsize{±0.1} \\
\midrule
BeiT-B16 &48.3\scriptsize{±0.0} &52.6\scriptsize{±0.0} \\
\small{+ CFA / SHOT-IM} &\textbf{35.4\scriptsize{±0.0}}/37.6\scriptsize{±0.0} &\textbf{47.5\scriptsize{±0.0}}/49.1\scriptsize{±0.0} \\
BeiT-L16 &35.9\scriptsize{±0.0} &44.2\scriptsize{±0.0} \\
\small{+ CFA / SHOT-IM} &\textbf{26.0\scriptsize{±0.0}}/28.2\scriptsize{±0.0} &\textbf{39.9\scriptsize{±0.1}}/41.5\scriptsize{±0.0} \\
\bottomrule
\end{tabular}
\caption{Adaptation results based on several backbone networks. The evaluation metric of ImageNet-C is the averaged top-1 error over 15 corruption types with a severity level of 5. We use publicly available models that were already fine-tuned on ImageNet.}
\label{table_model_agnosticity}
\end{table}

\begin{table}[t]\centering
\scalebox{0.95}{
\begin{tabular}{lrrrrr}\toprule
Severity &Source &Tent &SHOT-IM &CFA \\\midrule
1 &16.8±0.0 &15.4±0.0 &15.8±0.0 &\textbf{15.3±0.0} \\
2 &20.3±0.0 &17.9±0.0 &18.4±0.0 &\textbf{17.5±0.0} \\
3 &22.5±0.0 &19.6±0.4 &19.9±0.0 &\textbf{18.7±0.0} \\
4 &27.2±0.0 &24.0±1.2 &23.2±0.0 &\textbf{21.5±0.0} \\
5 &35.9±0.0 &33.6±0.1 &28.2±0.0 &\textbf{26.0±0.0} \\
\midrule
Average &24.5±0.0 &22.1±0.3 &21.1±0.0 &\textbf{19.8±0.0} \\
\bottomrule
\end{tabular}
}
\caption{Top-1 error rate on ImageNet-C averaged across all the severity level and 15 corruption types. BeiT-L16 is used as a model.}
\label{table_beit_result}
\end{table}

\begin{table}[t]\centering
\scalebox{0.95}{
\begin{tabular}{lrrr}\toprule
Method &W/. Eq(\ref{equ_h_std_s})(\ref{equ_h_std_t}) &W/O. Eq(\ref{equ_h_std_s})(\ref{equ_h_std_t}) \\\midrule
Source &61.95±0.00 &61.95±0.00 \\
\midrule
CFA-F ($K$=1) &46.69±0.02 &46.69±0.01 \\
CFA-F ($K$=3) &46.66±0.02 &47.28±0.03 \\
CFA-F ($K$=5) &46.64±0.02 &54.51±0.14 \\
\midrule
CFA-C &\textbf{45.31±0.03} &47.13±0.06 \\
\midrule
CFA ($K$=1) &43.98±0.04 &45.28±0.02 \\
CFA ($K$=3) &\textbf{43.90±0.04} &44.56±0.03 \\
CFA ($K$=5) &\textbf{43.90±0.04} &52.25±0.11 \\
\bottomrule
\end{tabular}
}
\caption{Ablation study of CFA. Top-1 error on ImageNet-C averaged over 15 corruption types with severity level of 5. ViT-B16 is used. CFA-F : Overall distribution matching only. CFA-C :  Class-conditional distribution matching only. \kojima{K : Maximum \# of central moments. K=1 denotes first-order moment (mean) matching only.}}
\label{table_ablation_study}
\end{table}

\subsection{Experiment Result}
\label{sec_quantitative_result}

\paragraph{Modulation Study.} 
Table \ref{table_modulation_study} answers the question about which set of modulation parameters is the most suitable for improving the performance of test-time adaptations on ViTs. There are two findings. First, updating layer nomalization parameters can achieve  balanced and high performance across all the main methods. Second, SHOT-IM and CFA achieve higher performance by updating all or feature extractor parameters, while Tent and PL deteriorates the performance because of catastrophic failure 
(See Appendix I for details). 
This indicates that a method with more sophisticated strategy within the adaptation function can work properly without sensibly selecting modulation parameters. In all the subsequent experiments, we choose layer normalization as modulation parameters across all the methods for the fair comparison.

\paragraph{CFA Outperforms Existing Methods on Several Datasets.}

Table \ref{table_method_comparison} summarizes the adaptation result across datasets for each test time adaptation methods. As for CIFAR-10-C, CIFAR-100-C, and ImageNet-C, we measure the averaged top-1 error across 15 corruption types for the highest severity level (=5). 
\kojima{CFA (our method) aligns both the overall distribution and class-conditional distribution between source and target datasets. In addition to CFA, we have experimented class-conditional distribution matching only method (CFA-C) and overall distribution matching only method (CFA-F) to measure the contribution of each distribution matching to performance. Specifically, the objective function of CFA-C and CFA-F is respectively defined as Eq.(\ref{equ_c_loss}) and Eq.(\ref{equ_f_loss}).}
The experiment results demonstrate that CFA can achieve the best or comparable performance against baseline methods across all datasets.
\kojima{
It is also verified that CFA-F and CFA-C can solely achieve better performance compared to the case without adaptation (``Source'') as in Table \ref{table_method_comparison}. Finally, CFA further boosts the performance on most datasets by combining them.}

\paragraph{CFA Is Model Agnostic.}
Table \ref{table_model_agnosticity} shows the adaptation results on ImageNet-C and ImageNet-Sketch by CFA (and SHOT-IM for comparison) based on various category's backbone networks. 
Specifically, we used publicly available models that are already fine-tuned on ImageNet-2012 at a resolution of 224 $\times$ 224, including ResNet, ViT, ViT-AugReg, DeiT, BeiT, and MLP-Mixer.
See Appendix D for details. 
The modulation parameters are BN for ResNet, and LN for the others. The results indicate that our method (CFA) consistently improves the performance regardless of the backbone networks. It is also found that the better performance on the source dataset (ImageNet), the stronger robustness on the target dataset (ImageNet-C) the model can gain by adaptation. 
See Appendix E for visualization of the relationship.

\paragraph{CFA Achieves SOTA Performance.}
Among these backbone networks, we select BeiT-L16, which achieved strong performance on ImageNet, and calculate the top-1 error rate on ImageNet-C averaged over 15 types of corruptions and all the severity levels (1-5) for each TTA methods. The results described in Table \ref{table_beit_result} demonstrate that 19.8\% using CFA on BeiT-L16 gives superior performance to the other baseline methods. It also outperforms the existing test-time adaptation result 44.0\% using Tent on ResNet50 \cite{wang2020tent}. Therefore, CFA achieves the state-of-the-art (SOTA) performance \kojima{among TTA methods that do not need to alter training phase} 
(See Appendix I for the full results).

\paragraph{Ablation Study.}

Table \ref{table_ablation_study} summarizes the ablation study results to analyze the detailed contributions of each components in our method on the robustness. \kojima{Specifically, we analyze the effect of the normalization of hidden representation before calculating the distribution statistics by comparing the scenarios of with/without Eq.(\ref{equ_h_std_s}) and (\ref{equ_h_std_t}).} 
In the case of CFA-F, it is verified that the performance deteriorates significantly without Eq.(\ref{equ_h_std_s}) and  (\ref{equ_h_std_t}) \kojima{especially when the maximum number of moments K gets larger. This indicates that feature normalization, especially bounding minimum and maximum value of hidden representation, stabilizes the performance of matching higher order moments. In the case of CFA-C, it is verified that the performance deteriorates without Eq.(\ref{equ_h_std_s}) and (\ref{equ_h_std_t}). It is speculated that feature normalization, especially layer normalization without affine transformation, might have a positive effect on class-conditional (centroid) distribution matching by highlighting the distribution property for each class}.
In addition, it is also verified that using both overall feature alignment and class-conditional feature alignment (CFA) boosts the performance compared to either alone (CFA-F or CFA-C) \kojima{regardless of the value of K}.

\paragraph{Hyperparameter Sensitivity.} 

\begin{figure}[t]
\centering
\includegraphics[width=1.0\columnwidth]{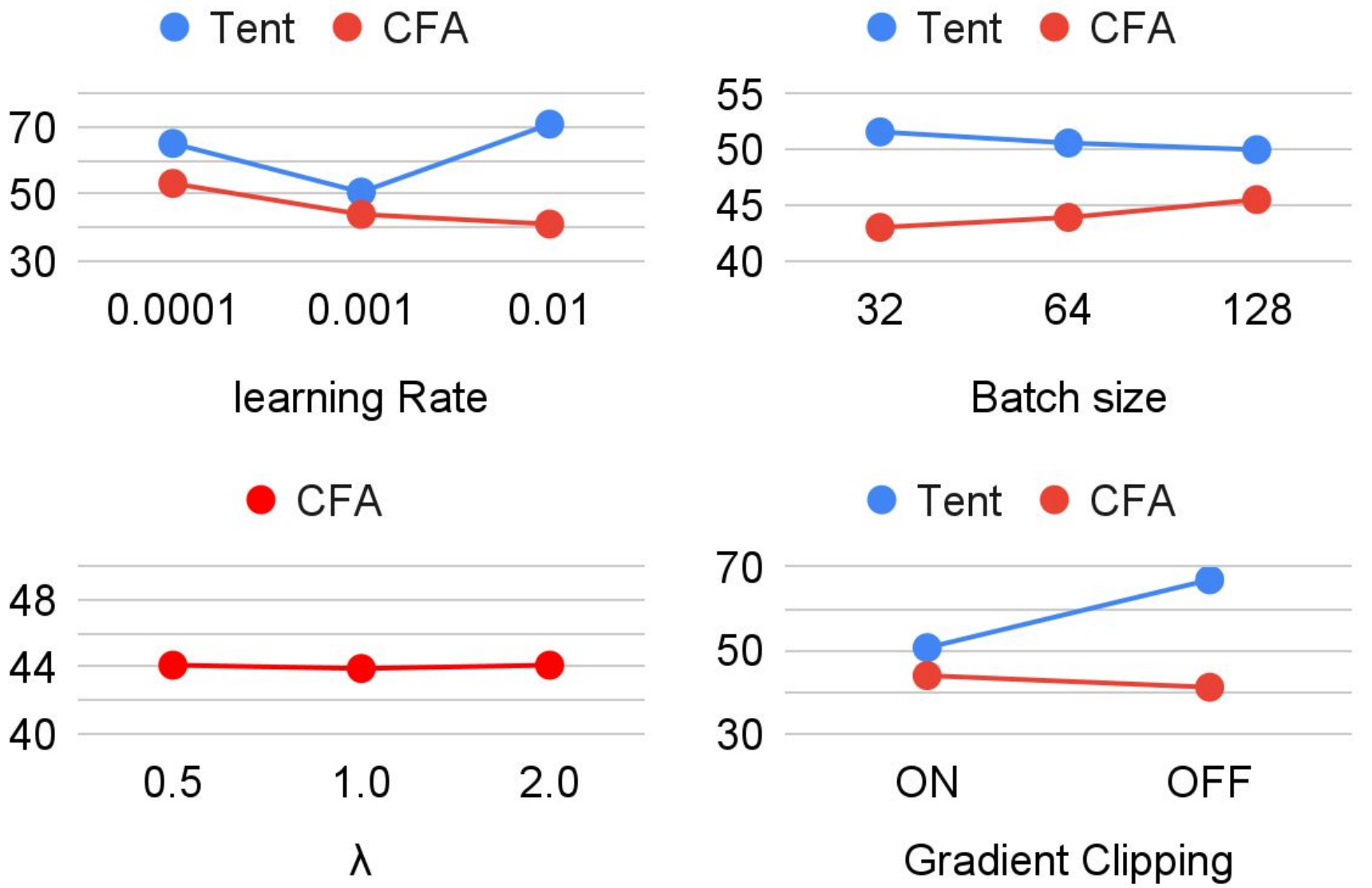}
\caption{The effect of changing hyperparameters on Tent and CFA performance. The evaluation metric is the top-1 error on ImageNet-C averaged over 15 corruption types with a severity level of 5. ViT-B16 is used as a model. Either one of the hyperparameter values is changed from the default described in \S \ref{sec_adapt_setting}.}
\label{fig_hypara}
\end{figure}

For online adapatation, hyperparameter selection is a challenging issue.  Figure \ref{fig_hypara} shows the experiment results about each hyperparameter sensitivity on ImageNet-C with the highest severity level (=5) averaged over 15 corruption types. We checked 4 hyperparameters by changing either one of the values from the default described in \S\ref{sec_adapt_setting}. (a) learning rate, (b) batch size, (c) balancing hyperparameter $\lambda$, and (d) whether to enable gradient clipping for SGD optimization. 
The finding is that Tent is more sensitive to some hyperparameters than CFA. In particular, enabling gradient clipping is essential when applying Tent to ViT to avoid catastrophic failure, while it is not essential for CFA. Furthermore, large learning rate also causes Tent catastrophic failure. In contrast, CFA is robust to all the above hyperparameters. This indicates that we can safely use CFA in unknown environments with rough hyperparameter selection.

\section{Conclusion}
This is the first study that verifies the effectiveness of test-time adaptation methods on \lm to boost their robustness.
Experiment results demonstrate that the existing methods can be applied to ViT and the prior-convention (sensibly selecting modulation parameters) is not necessary when a proper loss function is used. This study also proposed a novel method, CFA, which is hyperparameter friendly, model agnostic, and surpasses existing baselines. 
We hope this study becomes a milestone  \kojima{of TTA for current large models and will serve as a stepping stone to TTA for larger models in the future}.

\section*{Acknowledgements}

\kojima{
This work has been supported by the Mohammed bin Salman Center for Future Science and Technology for Saudi-Japan Vision 2030 at The University of Tokyo (MbSC2030). Computational resource of AI Bridging Cloud Infrastructure (ABCI) provided by National Institute of Advanced Industrial Science and Technology (AIST) was used for experiments.
}

\fontsize{9.5pt}{0.30cm}\selectfont
\bibliographystyle{named}
\nocite{deng2009imagenet}
\nocite{rusak2020simple}
\bibliography{ijcai22}
\normalsize

\include{ijcai22_appendix}

\end{document}

%% file: ijcai22_appendix.tex

\appendix
\section{Datasets Description}
\label{appendix_datasets}

\subsection{CIFAR-10/100 and ImageNet}

CIFAR-10 / CIFAR-100 and ImageNet are used as source datasets for our experiments. CIFAR-10 / CIFAR-100 are respectively 10-class / 100-class color image datasets including 50,000 training data and 10,000 test data with a resolution of $32{\times}32$. ImageNet is a 1000-class image dataset with more than 1.2 million training data and 50,000 validation data with various resolutions.

\subsection{Corruption Datasets}

\begin{figure*}[h]
\captionsetup{width=.80\linewidth}
\centering
\includegraphics[width=0.80\textwidth]{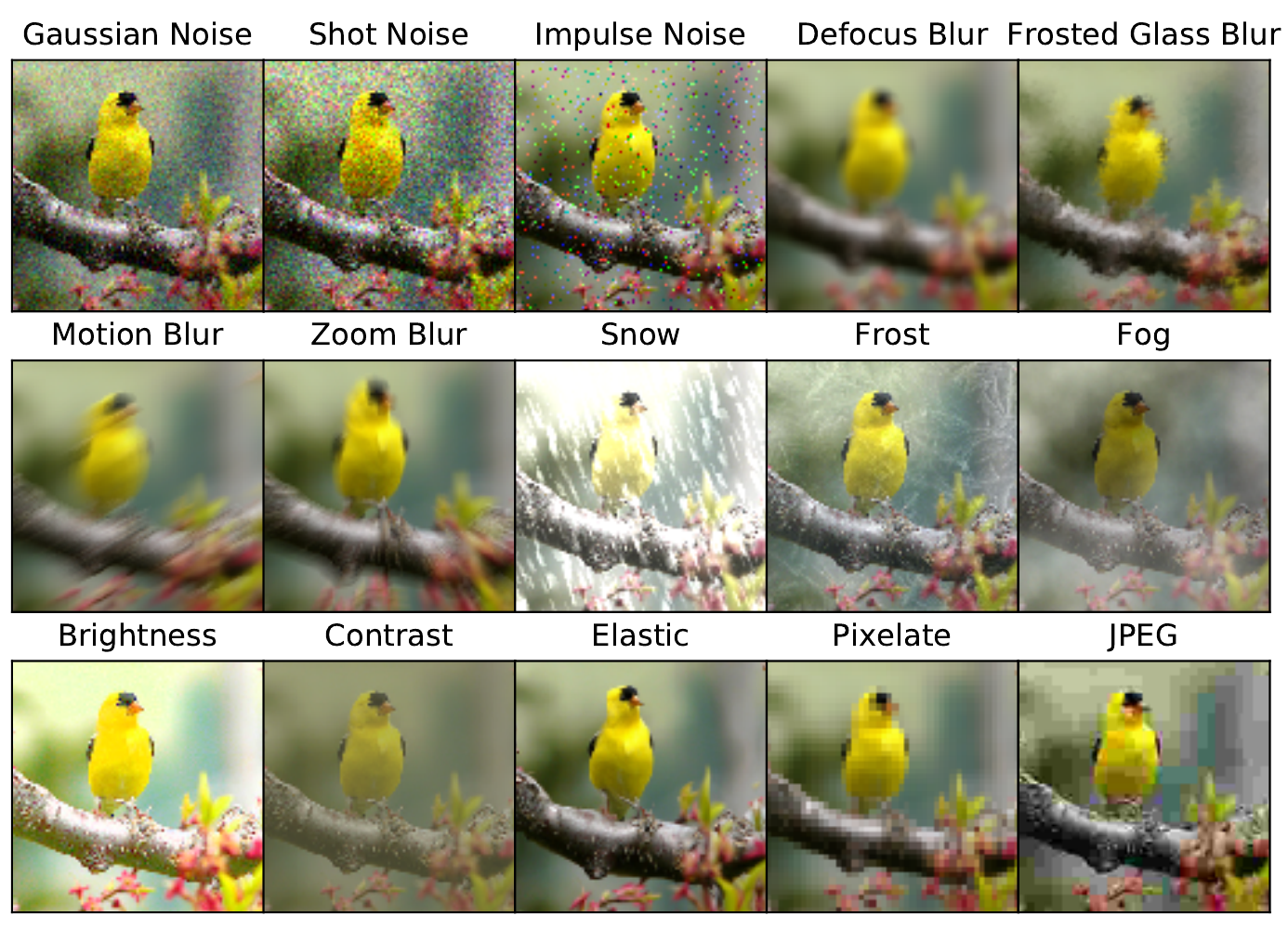}
\caption{Examples of 15 crruption types from ImageNet-C. These images are borrowed from
[Hendrycks and Dietterich, 2019]. CIFAR-10-C and CIFAR-100-C have the same corruption types.}
\label{fig_corruption_types}
\end{figure*}

\begin{figure*}[h]
\captionsetup{width=.80\linewidth}
\centering
\includegraphics[width=0.80\textwidth]{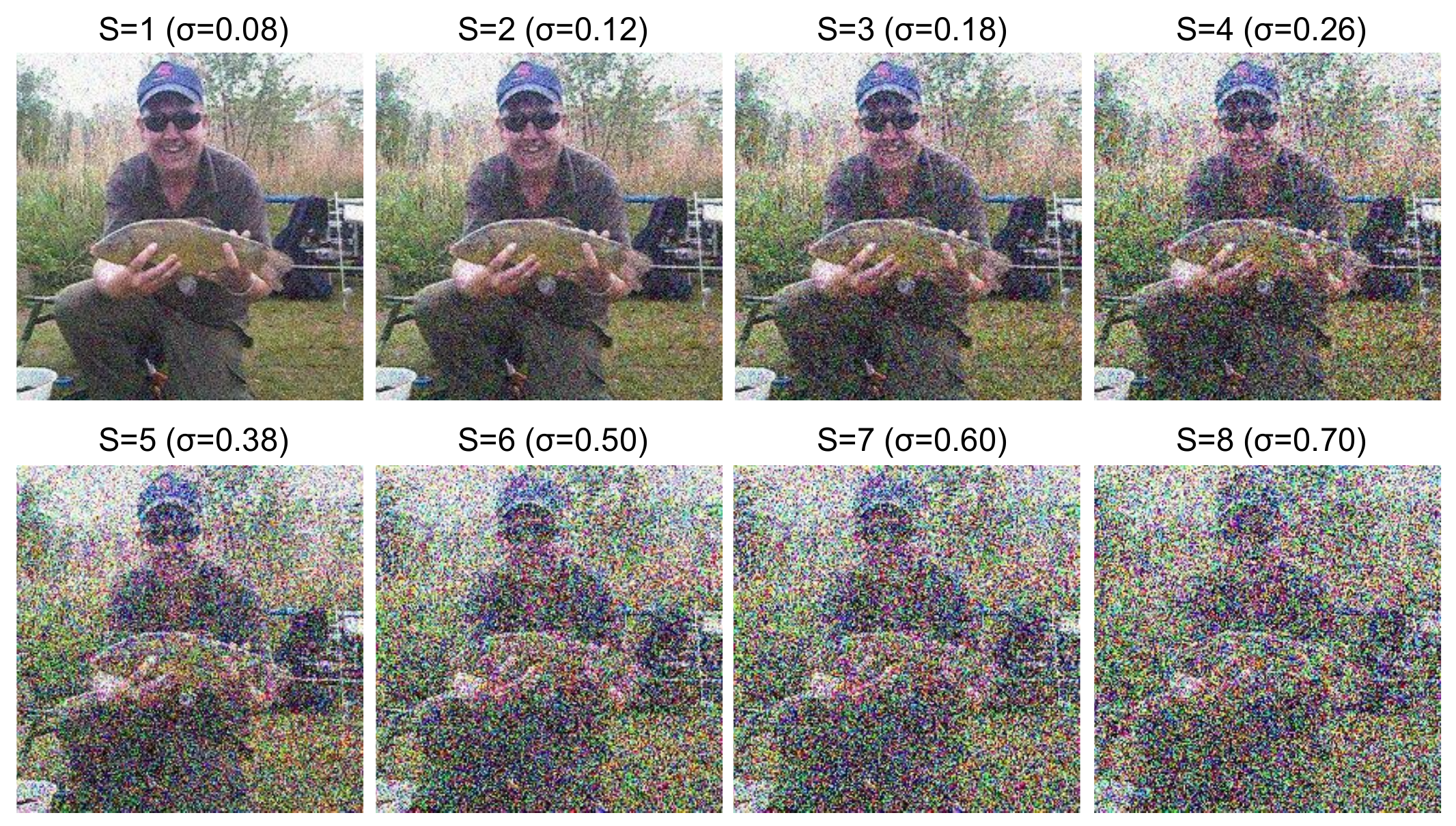}
\caption{Example images of different severity levels. S : Severity Level. $\sigma$ : standard deviation of Gaussian Noise. Images with severity level 1-5 are from ImageNet-C. Images with severity level 6-8 are created by adding Gaussian noise to the original ImageNet validation dataset.}
\label{fig_corruption_severity_levels}
\end{figure*}

CIFAR-10-C / CIFAR-100-C and ImageNet-C \cite{hendrycks2019robustness} are used as target datasets for our experiment. These datasets contain data with 15 types of corruptions with five levels of severity. Therefore, each dataset has 75 varieties of corruptions in total. Each corrupted data is composed of data from original CIFAR-10 and CIFAR-100 test images, and ImageNet validation images. Therfore, the CIFAR-10-C/CIFAR-100-C consists of 10,000 images for each corruption/type and ImageNet-C dataset consists of 50,000 images for each corruption/type.
Fig.\ref{fig_corruption_types} illustrates examples of 15 corruption types from ImageNet-C. These examples are cited from \cite{hendrycks2019robustness}. CIFAR-10-C and CIFAR-100-C have the same corruption types.
Figure.\ref{fig_corruption_severity_levels} illustrates examples for each corruption severity levels from 1 to 8 of Gaussian Noise. ImageNet-C dataset contains only images of severity level from 1 to 5. We have augmented the severity levels of Gaussian Noise datasets from 6 to 8 following the same setting as \cite{rusak2020simple} to use them for the corruption severity experiment on Table.\ref{table_severity}.

\subsection{Digits Datasets}

\begin{figure*}[h]
\captionsetup{width=.80\linewidth}
\centering
\includegraphics[width=0.80\textwidth]{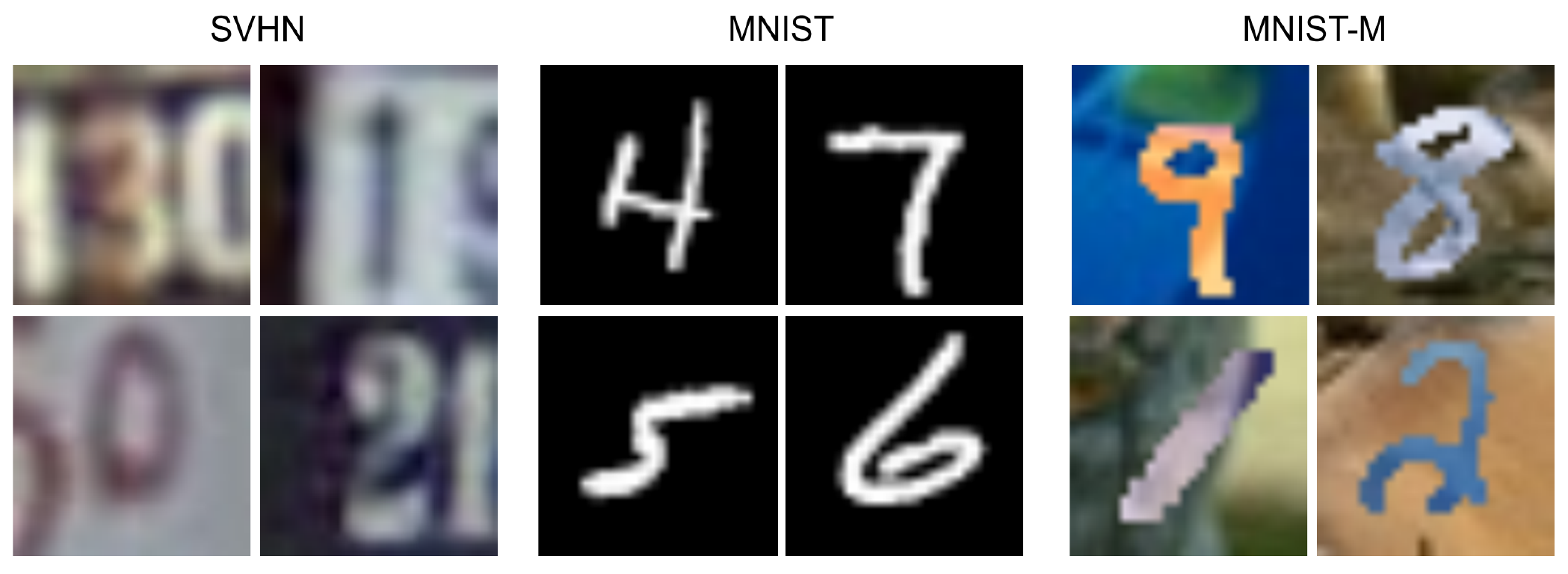}
\caption{Examples from Digit recognition Datasets.}
\label{fig_digits_samples}
\end{figure*}

SVHN, MNIST and MNIST-M are 10-class classification datasets for digit recognition ranging from 0 to 9. SVHN\cite{svhn} is used as source dataset. The total number of training data is 73,257. Test data of MNIST\cite{lecun1998gradient} and MNIST-M\cite{ganin2015unsupervised} are used as target. The total numbers of test dataset used for adaptation are 10,000 and 10,000, respectively. MNIST data is a gray scale image, so we convert the image into an RGB scale as a data preprocessing.
Figure.\ref{fig_digits_samples} illustrates some image examples from SVHN, MNIST and MNIST-M.

\subsection{ImageNet-Sketch}

\begin{figure*}[h]
\captionsetup{width=.80\linewidth}
\centering
\includegraphics[width=0.80\textwidth]{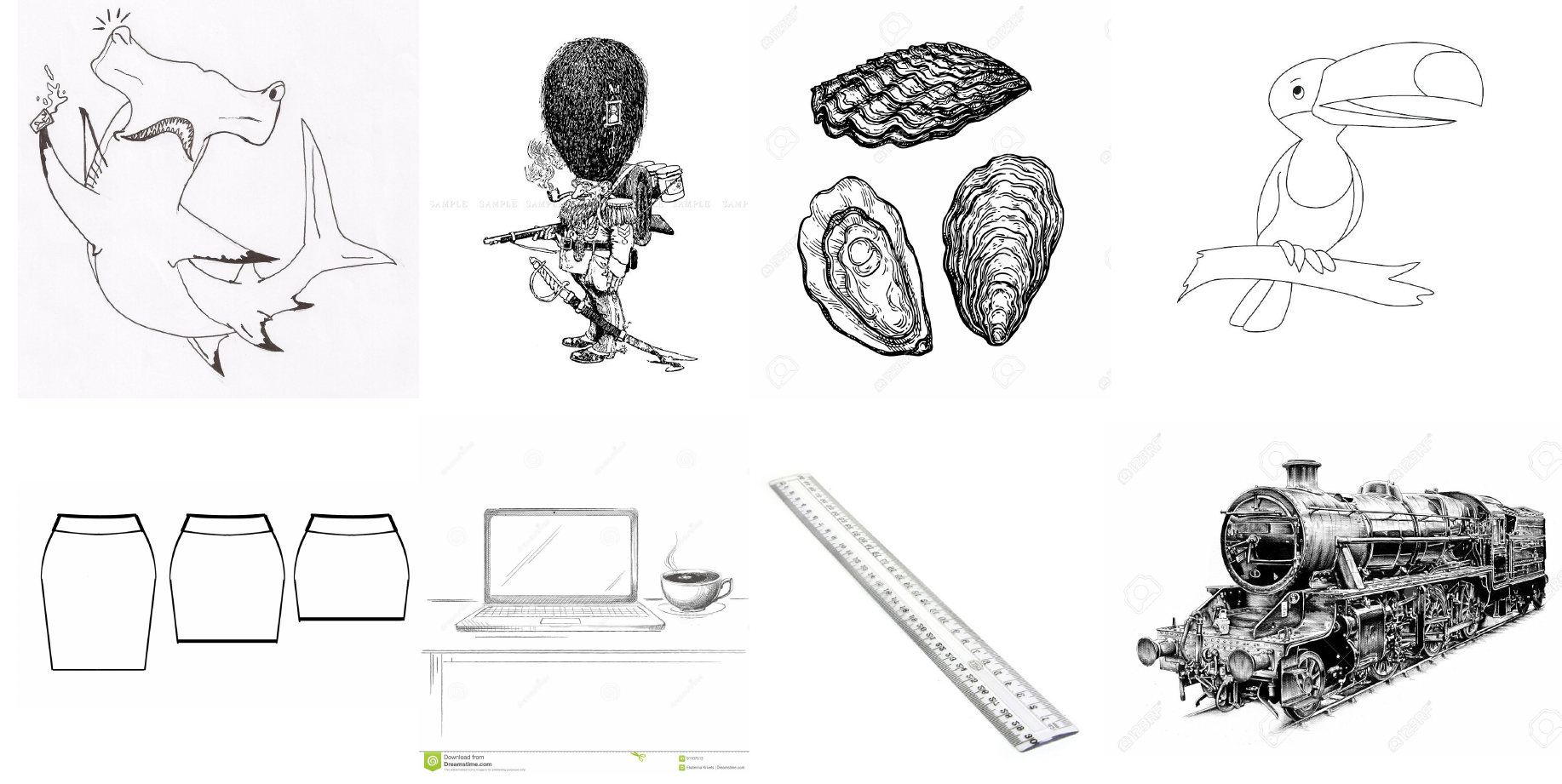}
\caption{Examples from ImageNet-Sketch Dataset.}
\label{fig_imagenet_sketch_samples}
\end{figure*}

ImageNet-Sketch \cite{wang2019learning} is used as target datasets for our experiments. ImageNet-Sketch contains 50,000 sketch images, 50 images for each of the 1,000 ImageNet classes. Figure.\ref{fig_imagenet_sketch_samples} illustrates image examples from ImageNet-Sketch. Most images are mono-color, but RGB resolution.

\section{Dataset Pre-processing}
\label{appendix_preprocessing}
For this experiment, images of all datasets are resized to $224{\times}224$. For ImageNet, some images are rectanglar, so all the images are \kojima{resized with fixed aspect ratio so that the shorter side of the rectangle is 256}, followed by center-cropping with a size of $224{\times}224$. ImageNet-C data have already been pre-processed in the same way and are publicly available. ImageNet-Sketch samples are resized to $224$ in the same way and center-cropped with a size of $224{\times}224$. When using ViT-B16, which is a baseline model for our experiment, pixels of images are rescaled from $[0, 255]$ to $[-1, 1]$ by taking the mean and std as $[0.5, 0.5, 0.5]$ and $[0.5, 0.5, 0.5]$ across all the datasets. See Table.\ref{table_mean_std_detail} for the other models.

\section{Fine-Tuning Details}
\label{appendix_finetuning}
Vistion Transformer(ViT) is used as a basic model for this experiment. Before adaptation, the model needs to be fine-tuned for each dataset. For CIFAR-10, CIFAR-100 and SVHN, we use ViT-B16 parameters that is already pre-trained on ImageNet-21K\cite{deng2009imagenet},  which is a large dataset with 21k classes and 14M images . For fine-tuning hyperparameters, following \cite{dosovitskiy2020image}, We use batch size of 512, set optimizer as SGD with momentum 0.9 and gradient clipping at global norm 1.0. We choose a learning rate of 0.03. We apply a cosine schedule of 500 warmup steps and the total number of iterations as 2000 for CIFAR-100 and SVHN. We apply cosine schedule of 200 warmup steps  and the total number of iteration as 1000 for CIFAR-10.  The fine-tuning result is 1.1\% Top-1 error for CIFAR-10, 6.8\% for CIFAR-100, and 3.0\% for SVHN. For ImageNet, we use publicly available ViT-B16 parameters that are already pre-trained for ImageNet-21K and fine-tuned for ImageNet(-2012). Therefore, fine-tuning on ImageNet is not required. The Top-1 error for ImageNet is 18.6\% in our setting. 

\section{Detailed information about the backbone networks used in Table \ref{table_model_agnosticity}}
\label{appendix_models}

Table.\ref{table_backbone_detail} summarizes the detailed information about the backbone networks. Note that all the models used are already pre-trained on ImageNet-2012 at resolution 224 $\times$ 224 and publicly available.For model agnostic study, pixels are rescaled by taking the mean and std as described at Table \ref{table_mean_std_detail}.

\section{Visualization of Model Agnosticity Study}
\label{appendix_model_agnostic_visualization}

\begin{figure*}[h]
\captionsetup{width=.70\linewidth}
\centering
\includegraphics[width=0.70\textwidth]{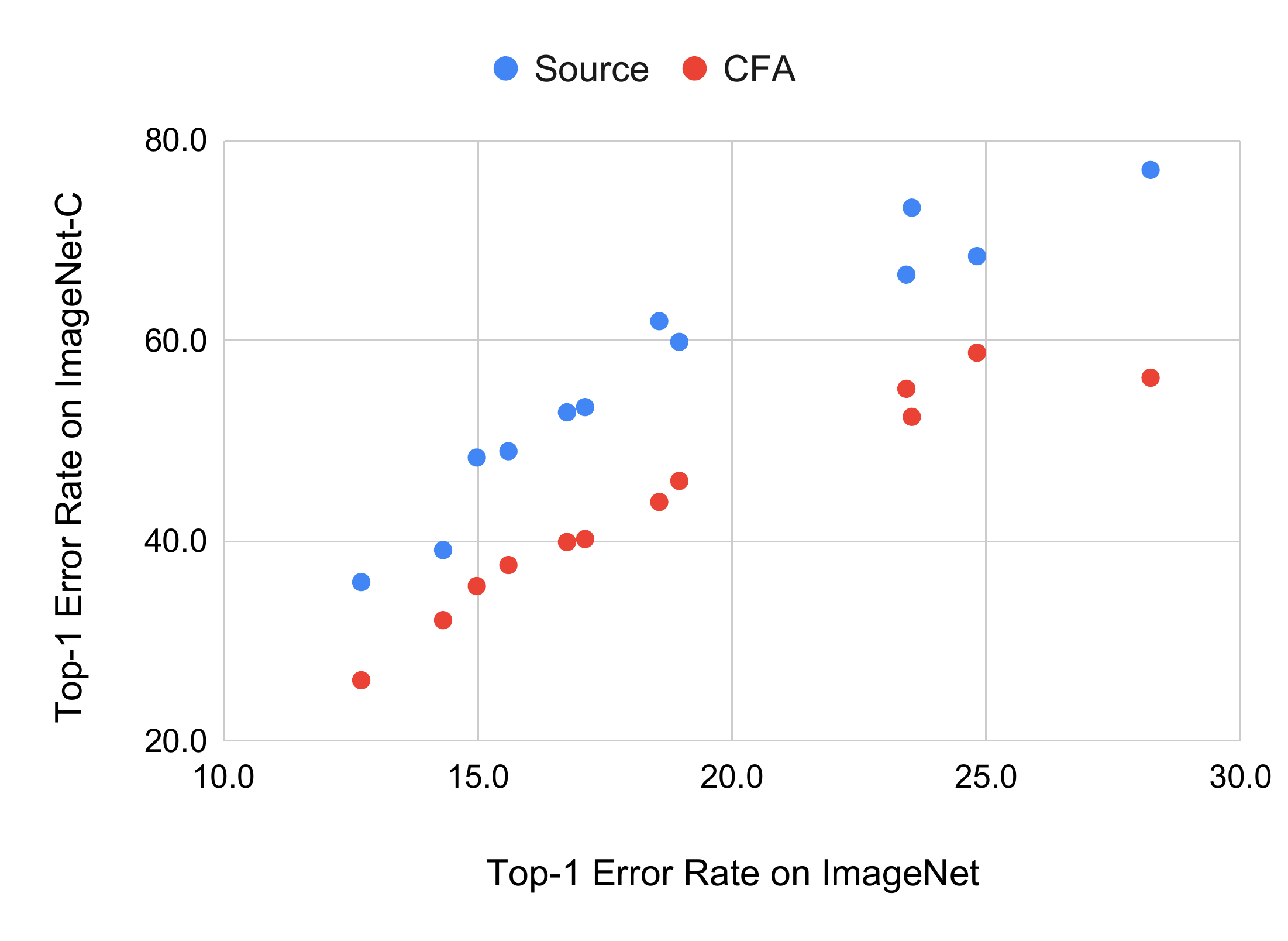}
\caption{Visualization of model agnosticity study result. Each plot indicates the relationship between the Top-1 error rate on ImageNet (Source dataset) and the Top-1 error rate on ImageNet-C (Target dataset) before / after the CFA adaptation for each backbone network.}
\label{fig_model_agnostic_visualization}
\end{figure*}

Figure \ref{fig_model_agnostic_visualization} visualizes the relationship between the Top-1 error rate on ImageNet (Source dataset) and the Top-1 error rate on ImageNet-C (Target dataset) before / after the CFA adaptation for each backbone network.

\section{Detail Settings of Baseline Methods}
\label{appendix_detail_baseline}

\subsection{TFA(-)} Test-time feature alignment (TFA) \cite{liu2021ttt} aligns the hidden representation on target data by minimizing the distance of the mean vector $\mu^s, \mu^t \in \mathbb{R}^D$ and covariance matrix $\Sigma^s, \Sigma^t \in \mathbb{R}^{D \times D}$ between source and target. $D$ is the dimension size of the hidden representation. We focus only on the ''Online Feature Alignment'' part in TTT++ \cite{liu2021ttt}. Original TFA \cite{liu2021ttt} aligns the distributions at both the hidden representation and the output of the self-supervised head. However, in our experiment, TFA(-) does not employ self-supervised learning, so we only focus on distribution matching of the hidden representation. Specifically, in this experiment, the hidden representation to align is defined as the one before the classifier head $h(x) = f(x;\phi)$. The loss function is $\mathcal{L} = \beta_1 ||\mu^s - \mu^t ||^2_2 + \beta_2 || \Sigma^s - \Sigma^t ||^2_F$ where $||\cdot||_2$ is the Euclidean norm and $||\cdot||_F$ is the Frobenius norm. $\beta_1$ and $\beta_2$ are balancing hyperparameters. Like CFA, TFA(-) calulates the statistics on source dataset and store them in memory before adaptation. Note that ''Online Dynamic Queue'' \cite{liu2021ttt} is not used in TFA(-) in our experiment. Table \ref{table_TFA_experiment_detail} describes the small experiment results of TFA(-) on ImageNet-C datasets with severity=5 by changing the balancing hyperparameters $\beta_1, \beta_2$. Following \cite{liu2021ttt}, we use $\beta_1=1, \beta_2=1$ for the main experiment in Table \ref{table_method_comparison}.

\begin{table}[h]\centering
\begin{tabular}{lrr}\toprule
Method&ImageNet-C \\\midrule
TFA(-) ($\beta_1=1, \beta_2=1$) &57.7±0.1 \\
TFA(-) ($\beta_1=1, \beta_2=1/D$) &48.8±0.0 \\
TFA(-) ($\beta_1=1, \beta_2=0$) &46.7±0.0 \\
TFA(-) ($\beta_1=0, \beta_2=1/D$) &65.5±0.4 \\
\midrule
TFA(-) ($\beta_1=1/D, \beta_2=1/D$) &61.0±0.1 \\
TFA(-) ($\beta_1=1/D, \beta_2=1/D^2$) &51.8±0.0 \\
TFA(-) ($\beta_1=1/D, \beta_2=0$) &51.8±0.0 \\
TFA(-) ($\beta_1=0, \beta_2=1/D^2$) &62.0±0.0 \\
\bottomrule
\end{tabular}
\caption{Experiment results of TFA(-) on ImageNet-C by changing the balancing hyperparameters $\beta_1, \beta_2$. Evaluation metric is Top-1 error on ImageNet-C averaged over 15 corruption types with severity level=5. ViT-B16 is used as a model.}
\label{table_TFA_experiment_detail}
\end{table}

\subsection{T3A} 
T3A \cite{iwasawa2021testtime} updates the centroid of each class averaging over the pseudo labeled samples’ feature vectors in an online manner. This is a gradient-free approach and there is no loss function. The hyperparameter filter size K is set to 100 in our experiment. 

\section{Central Moment Distance (CMD)}
\label{appendix_cmd}

Our proposal utilizes central moment distance (CMD) \cite{CMD} as to minimize the overall distribution differences between the source and target data. CMD is an existing approach for UDA settings. Formally, let $X$ and $Y$ be bounded random samples with respective probability
distributions $p$ and $q$ on the interval $[a; b]^N$. 
CMD is defined by
\begin{equation}
\begin{split} 
  CMD_k &= \frac{1}{|b-a|} || \mathbb{E}(X) - \mathbb{E}(Y) ||_2 
  \\ &+ \frac{1}{|b-a|^k} \sum_{k=2}^K || \mathbb{M}_k(X) - \mathbb{M}_k(Y) ||_2 ,
\end{split}
\label{equ_cmd}
\end{equation}
where $\mathbb{E}(X) = \frac{1}{|X|}\sum_{x}$ is the empirical expectation vector computed for the sample $X$, and $\mathbb{M}_k(X) = \mathbb{E}((x-\mathbb{E}(X))^k)$ is the vector of all $k^{th}$ order central moments of the coordinates of $X$. $Y$ follows the same idea. 
Previous studies have focused on using CMD in the field of UDA to reduce the distributional gaps between source and target representations. However, CMD can also be potentially used for test-time adaptation because it does not need to store the source dataset itself; instead, we store central moment statistics of the source data in memory, and use it during online adaptation for moment matching between source and target. 

\section{Further Study on Severer Corruption}
\label{appendix_severer_corruption}

We further analyse the robustness of Tent and CFA from the severity viewpoint. The detail experiment setting is the same as described at \S\ref{sec_adapt_setting}. Table \ref{table_severity} summarizes the results on Gaussian Noise for each severity level from 1 to 8. 
Original ImageNet-C contains only severity level of until 5, so we create the severer corrupted images (severity 6, 7 and 8) specific to Gaussian Noise by increasing the standard deviation. We found that gradient clipping\cite{Zhang2020Why} is essential to use Tent in ViT, which are not used in the original study. Specifically, we clip the global norm of gradients to $1.0$. Without gradient clipping, Tent often gave catastrophic failure. However, even with gradient clipping, when the noise is severer, Tent causes catastrophic failure while CFA avoids it during adaptation.

\begin{table}[t]\centering
\begin{tabular}{lrrrrr}\toprule
Severity &Source &Tent &Tent &CFA \\
(std) & &w/o GC &w/ GC &w/ GC \\\midrule
1 (0.08) &23.3±0.0 &\textbf{22.0±0.1} &22.0±0.0 &22.1±0.1 \\
2 (0.12) &26.9±0.0 &\textbf{24.8±0.0} &24.9±0.1 &25.0±0.0 \\
3 (0.18) &35.6±0.0 &\textbf{30.5±0.1} &31.1±0.1 &30.6±0.1 \\
4 (0.26) &52.0±0.0 &59.8±23.0 &41.6±0.1 &\textbf{40.0±0.0} \\
5 (0.38) &77.7±0.0 &97.3±0.3 &66.3±8.8 &\textbf{56.5±0.1} \\
6 (0.50) &92.7±0.0 &99.8±0.1 &99.1±0.3 &\textbf{72.0±0.1} \\
7 (0.60) &97.4±0.0 &99.9±0.0 &99.8±0.0 &\textbf{82.4±0.1} \\
8 (0.70) &99.0±0.0 &99.9±0.0 &99.9±0.0 &\textbf{89.8±0.2} \\
\bottomrule
\end{tabular}
\caption{Top-1 error based on Gaussian Noise Dataset for each severity. ViT-B16 is used as a model. GC : Gradient Clipping. $\sigma$ : Standard deviation of Gaussian noise. We additionally create severity level 6-8 by adding the noise to the ImageNet validation data.}
\label{table_severity}
\end{table}

\section{Detail Experiment Results}
\label{appendix_detail_experiment}

\subsection{Detail Results for Table.\ref{table_modulation_study}}

See Table.\ref{table_modulation_study_detail_vitB16} and \ref{table_modulation_study_detail_vitL16}.

\subsection{Detail Results for Table.\ref{table_method_comparison}}

See Table.\ref{table_method_comparison_detail}.

\subsection{Detail Results for Table.\ref{table_model_agnosticity}}

See Table.\ref{table_model_agnosticity_detail_1}, \ref{table_model_agnosticity_detail_2} and \ref{table_model_agnosticity_detail_3}.

\subsection{Detail Results for Table.\ref{table_beit_result}}

See Table.\ref{table_beit_result_detail}.

\subsection{Detail Results for Table.\ref{table_ablation_study}}

See Table.\ref{table_ablation_study_detail}.

\subsection{Detail Results for Figure.\ref{fig_hypara}}

See Table.\ref{table_hypara_detail}.


\begin{table*}[t]\centering
\begin{tabular}{lrrrr}\toprule
Backbone Networks &Pytorch &Model Name &ImageNet Top-1 \\
&Library(Ver.) &in Library &Error Rate \\\midrule
ResNet50 &torchvision(0.10.0) &resnet50 &24.8 \\
ResNet101 &torchvision(0.10.0) &resnet101 &23.4 \\
ViT-B16 &timm(0.4.9) &vit\_base\_patch16\_224 &18.6 \\
ViT-L16 &timm(0.4.9) &vit\_large\_patch16\_224 &17.1 \\
DeiT-S16 &timm(0.5.0) &deit\_small\_distilled\_patch16\_224 &19.0 \\
DeiT-B16 &timm(0.5.0) &deit\_base\_distilled\_patch16\_224 &16.8 \\
MLP-Mixer-B16 &timm(0.5.0) &mixer\_b16\_224 &23.5 \\
MLP-Mixer-L16 &timm(0.5.0) &mixer\_l16\_224 &28.2 \\
ViT-B16 (AugReg) &timm(0.5.0) &vit\_base\_patch16\_224 &15.6 \\
ViT-L16 (AugReg) &timm(0.5.0) &vit\_base\_patch16\_224 &14.3 \\
BeiT-B16 &timm(0.5.0) &beit\_base\_patch16\_224 &15.0 \\
BeiT-L16 &timm(0.5.0) &beit\_large\_patch16\_224 &12.7 \\
\bottomrule
\end{tabular}
\caption{Backbone networks information used in Table \ref{table_model_agnosticity}}
\label{table_backbone_detail}
\end{table*}

\begin{table*}[t]\centering
\captionsetup{width=.60\linewidth}
\begin{tabular}{lrrr}\toprule
Backbone Network &MEAN &STD \\\midrule
ResNet50 &[0.485, 0.456, 0.406] &[0.229, 0.224, 0.225] \\
ResNet101 &[0.485, 0.456, 0.406] &[0.229, 0.224, 0.225] \\
ViT-B16 &[0.5, 0.5, 0.5] &[0.5, 0.5, 0.5] \\
ViT-L16 &[0.5, 0.5, 0.5] &[0.5, 0.5, 0.5] \\
DeiT-S16 &[0.485, 0.456, 0.406] &[0.229, 0.224, 0.225] \\
DeiT-B16 &[0.485, 0.456, 0.406] &[0.229, 0.224, 0.225] \\
MLP-Mixer-B16 &[0.5, 0.5, 0.5] &[0.5, 0.5, 0.5] \\
MLP-Mixer-L16 &[0.5, 0.5, 0.5] &[0.5, 0.5, 0.5] \\
ViT-B16 (AugReg) &[0.5, 0.5, 0.5] &[0.5, 0.5, 0.5] \\
ViT-L16 (AugReg) &[0.5, 0.5, 0.5] &[0.5, 0.5, 0.5] \\
BeiT-B16 &[0.5, 0.5, 0.5] &[0.5, 0.5, 0.5] \\
BeiT-L16 &[0.5, 0.5, 0.5] &[0.5, 0.5, 0.5] \\
\bottomrule
\end{tabular}
\caption{MEAN and STD used for data pre-processing on ImageNet, ImageNet-C and ImageNet-Sketch for each backbone network.}
\label{table_mean_std_detail}
\end{table*}

\begin{table*}[t]\centering
\begin{tabular}{lrrrrrrrrr}\toprule
Method(Params) &Gaussian &Shot &Impulse &Defocus &Glass &Motion &Zoom &Snow \\ 
\midrule
Source &77.7±0.0 &75.1±0.0 &77.0±0.0 &66.9±0.0 &69.1±0.0 &58.5±0.0 &62.8±0.0 &60.9±0.0 \\
Tent(LN) &66.3±8.8 &77.8±1.6 &59.3±0.2 &50.9±0.2 &49.8±0.2 &46.7±0.0 &50.9±0.2 &75.7±2.0 \\
Tent(CLS) &74.9±0.1 &72.6±0.1 &74.4±0.0 &63.7±0.2 &66.8±0.3 &55.8±0.0 &61.0±0.0 &60.5±0.0 \\
Tent(Feature) &77.0±5.6 &87.5±6.0 &78.8±3.7 &48.8±0.1 &47.6±0.1 &44.5±0.1 &67.5±4.3 &84.6±4.4 \\
Tent(ALL) &78.2±7.2 &87.8±5.4 &79.7±4.6 &49.0±0.1 &47.7±0.2 &44.6±0.1 &70.9±4.0 &86.7±4.1 \\
PL(LN) &75.5±8.3 &71.7±3.7 &73.7±8.9 &53.2±0.1 &52.4±0.1 &48.6±0.3 &52.4±0.4 &75.3±1.8 \\
PL(CLS) &76.3±0.0 &74.0±0.1 &75.6±0.0 &66.4±0.0 &68.7±0.0 &57.0±0.1 &61.8±0.2 &60.7±0.0 \\
PL(Feature) &85.5±7.3 &85.6±8.7 &80.4±9.9 &51.6±0.2 &50.9±0.1 &46.9±0.0 &61.3±2.1 &88.7±2.5 \\
PL(ALL) &83.6±11.9 &90.7±2.8 &79.1±13.5 &51.7±0.4 &51.1±0.4 &46.8±0.3 &61.1±2.5 &88.8±2.3 \\
SHOT-IM(LN) &58.6±0.1 &56.4±0.1 &57.6±0.1 &49.7±0.1 &48.7±0.2 &46.0±0.1 &46.4±0.1 &47.0±0.1 \\
SHOT-IM(CLS) &75.5±0.0 &73.1±0.0 &74.9±0.0 &64.5±0.2 &68.4±0.0 &56.5±0.1 &61.4±0.1 &60.6±0.0 \\
SHOT-IM(Feature) &57.8±0.2 &55.3±0.1 &57.1±0.1 &47.8±0.1 &46.7±0.1 &43.7±0.1 &44.1±0.2 &44.4±0.2 \\
SHOT-IM(ALL) &57.8±0.1 &55.3±0.1 &57.1±0.2 &48.0±0.0 &46.8±0.1 &43.8±0.1 &44.2±0.1 &44.4±0.1 \\
CFA(LN) &56.5±0.1 &54.2±0.1 &55.4±0.1 &48.3±0.0 &47.1±0.0 &44.3±0.0 &44.4±0.2 &44.9±0.1 \\
CFA(CLS) &74.0±0.1 &71.3±0.1 &73.5±0.0 &60.9±0.2 &63.5±0.1 &54.9±0.1 &60.2±0.0 &60.0±0.0 \\
CFA(Feature) &55.1±0.2 &52.7±0.1 &54.2±0.1 &46.7±0.1 &44.9±0.1 &41.7±0.1 &41.5±0.1 &42.0±0.1 \\
CFA(ALL) &55.1±0.2 &52.7±0.1 &54.2±0.1 &46.7±0.1 &44.9±0.1 &41.7±0.1 &41.5±0.1 &42.0±0.1 \\
\midrule
Method(Params) &Frost &Fog &Brightness &Contrast &Elastic &Pixelate &Jpeg &Average \\
\midrule
Source &57.6±0.0 &62.9±0.0 &31.6±0.0 &88.9±0.0 &51.9±0.0 &45.3±0.0 &42.9±0.0 &61.9±0.0 \\
Tent(LN) &48.2±0.4 &44.3±0.3 &26.1±0.1 &58.5±0.3 &37.6±0.3 &32.7±0.1 &34.7±0.1 &50.6±0.5 \\
Tent(CLS) &57.2±0.0 &56.9±0.2 &31.5±0.0 &79.4±0.2 &51.8±0.0 &43.2±0.0 &42.1±0.1 &59.4±0.0 \\
Tent(Feature) &71.8±6.2 &53.2±19.7 &25.4±0.1 &54.9±0.3 &36.6±0.3 &30.8±0.0 &33.4±0.1 &56.2±2.2 \\
Tent(ALL) &71.9±3.8 &88.8±1.0 &25.4±0.0 &54.9±0.3 &37.0±0.4 &30.8±0.0 &33.5±0.1 &59.1±1.0 \\
PL(LN) &51.4±0.4 &79.5±3.3 &26.8±0.1 &62.8±0.2 &41.1±0.2 &35.1±0.2 &36.2±0.1 &55.7±1.4 \\
PL(CLS) &57.4±0.0 &59.0±0.1 &31.5±0.0 &81.9±0.2 &51.9±0.0 &44.5±0.1 &42.7±0.0 &60.6±0.0 \\
PL(Feature) &62.4±10.8 &91.6±0.5 &26.2±0.1 &71.0±10.4 &41.0±0.3 &33.2±0.2 &35.0±0.2 &60.8±2.1 \\
PL(ALL) &65.1±11.5 &93.8±0.1 &26.3±0.1 &72.4±11.6 &41.6±0.3 &33.3±0.1 &35.0±0.2 &61.4±2.2 \\
SHOT-IM(LN) &45.9±0.1 &43.2±0.2 &26.1±0.0 &56.9±0.2 &35.9±0.2 &32.4±0.1 &34.3±0.1 &45.7±0.0 \\
SHOT-IM(CLS) &57.3±0.0 &57.3±0.2 &31.5±0.0 &79.4±0.1 &51.8±0.0 &44.3±0.0 &42.6±0.0 &59.9±0.0 \\
SHOT-IM(Feature) &44.3±0.1 &40.3±0.1 &25.4±0.1 &53.2±0.1 &34.7±0.1 &30.7±0.1 &33.2±0.0 &43.9±0.0 \\
SHOT-IM(ALL) &44.3±0.1 &40.3±0.1 &25.4±0.0 &53.4±0.2 &34.8±0.1 &30.7±0.1 &33.3±0.1 &44.0±0.0 \\
CFA(LN) &44.8±0.1 &41.2±0.1 &25.6±0.1 &54.4±0.2 &33.2±0.1 &30.5±0.0 &33.5±0.1 &43.9±0.0 \\
CFA(CLS) &56.6±0.0 &56.4±0.1 &31.3±0.0 &78.0±0.1 &50.6±0.1 &41.2±0.1 &40.1±0.1 &58.2±0.0 \\
CFA(Feature) &42.5±0.1 &38.1±0.0 &24.8±0.1 &50.1±0.1 &31.8±0.0 &28.7±0.0 &32.2±0.0 &41.8±0.0 \\
CFA(ALL) &42.5±0.1 &38.1±0.0 &24.8±0.1 &50.1±0.1 &31.8±0.0 &28.7±0.0 &32.2±0.0 &41.8±0.0 \\
\bottomrule
\end{tabular}
\caption{Detailed experiment results of Table \ref{table_modulation_study} (Top-1 error rate) based on ViT-B16.}
\label{table_modulation_study_detail_vitB16}
\end{table*}

\begin{table*}[t]\centering
\begin{tabular}{lrrrrrrrrr}\toprule
Method(Params) &Gaussian &Shot &Impulse &Defocus &Glass &Motion &Zoom &Snow \\ 
\midrule
Source &60.5±0.0 &59.2±0.0 &59.1±0.0 &61.3±0.0 &62.6±0.0 &51.0±0.0 &54.2±0.0 &53.4±0.0 \\
Tent(LN) &48.9±0.1 &47.6±0.0 &48.0±0.1 &46.4±0.1 &48.2±0.1 &42.0±0.1 &44.6±0.1 &45.3±0.2 \\
Tent(CLS) &57.9±0.0 &56.5±0.0 &56.7±0.0 &55.3±0.3 &58.9±0.3 &46.5±0.1 &52.1±0.1 &51.5±0.1 \\
Tent(Feature) &47.7±0.1 &45.7±0.2 &46.8±0.2 &42.2±0.2 &43.2±0.3 &38.9±0.1 &40.3±0.3 &65.8±9.9 \\
Tent(ALL) &47.6±0.1 &45.8±0.2 &46.8±0.2 &42.3±0.2 &43.3±0.3 &39.0±0.1 &40.6±0.3 &70.9±5.3 \\
PL(LN) &50.4±0.0 &49.0±0.2 &49.2±0.2 &49.4±0.1 &51.6±0.3 &43.1±0.1 &46.1±0.1 &47.7±0.3 \\
PL(CLS) &58.9±0.1 &57.5±0.0 &57.7±0.1 &56.6±0.2 &61.3±0.1 &47.2±0.3 &52.4±0.1 &52.3±0.0 \\
PL(Feature) &48.6±0.2 &47.2±0.1 &48.0±0.1 &44.3±0.3 &46.0±0.4 &40.3±0.2 &43.0±1.1 &74.2±6.2 \\
PL(ALL) &48.5±0.3 &47.3±0.1 &47.9±0.1 &44.4±0.3 &46.2±0.1 &40.5±0.2 &43.7±1.5 &74.5±4.2 \\
SHOT-IM(LN) &48.8±0.0 &47.5±0.1 &47.9±0.1 &45.9±0.0 &47.6±0.1 &41.7±0.1 &43.7±0.1 &44.5±0.1 \\
SHOT-IM(CLS) &58.5±0.0 &57.2±0.0 &57.3±0.0 &55.6±0.5 &60.4±0.2 &46.8±0.1 &52.0±0.0 &52.0±0.1 \\
SHOT-IM(Feature) &47.3±0.1 &45.2±0.0 &46.3±0.1 &41.9±0.1 &42.5±0.2 &38.9±0.1 &39.3±0.1 &40.0±0.1 \\
SHOT-IM(ALL) &47.4±0.1 &45.3±0.1 &46.3±0.2 &41.9±0.1 &42.5±0.2 &38.9±0.1 &39.2±0.1 &40.0±0.1 \\
CFA(LN) &47.3±0.0 &45.9±0.1 &46.4±0.1 &44.6±0.1 &45.8±0.0 &40.3±0.1 &41.9±0.1 &42.2±0.1 \\
CFA(CLS) &56.7±0.1 &55.1±0.1 &55.5±0.0 &55.4±0.0 &56.3±0.3 &46.0±0.1 &51.1±0.0 &50.8±0.1 \\
CFA(Feature) &45.4±0.2 &43.5±0.1 &44.5±0.2 &40.7±0.2 &40.8±0.1 &37.1±0.2 &37.2±0.0 &37.6±0.1 \\
CFA(ALL) &45.4±0.2 &43.5±0.1 &44.5±0.2 &40.7±0.2 &40.8±0.1 &37.1±0.2 &37.2±0.0 &37.6±0.1 \\
\midrule
Method(Params) &Frost &Fog &Brightness &Contrast &Elastic &Pixelate &Jpeg &Average \\
\midrule
Source &52.6±0.0 &57.6±0.0 &28.3±0.0 &83.8±0.0 &45.3±0.0 &34.1±0.0 &37.5±0.0 &53.4±0.0 \\
Tent(LN) &46.3±0.2 &41.4±0.1 &24.5±0.2 &54.2±0.1 &37.8±0.2 &28.9±0.0 &30.9±0.1 &42.3±0.0 \\
Tent(CLS) &51.4±0.1 &50.5±0.1 &28.1±0.0 &75.1±0.1 &44.6±0.0 &33.5±0.0 &36.0±0.1 &50.3±0.0 \\
Tent(Feature) &80.1±2.7 &48.0±18.3 &23.2±0.0 &46.8±0.1 &32.5±0.3 &26.4±0.1 &29.0±0.2 &43.8±0.6 \\
Tent(ALL) &79.6±4.2 &49.4±20.6 &23.3±0.0 &46.7±0.1 &32.7±0.3 &26.4±0.1 &29.0±0.1 &44.2±1.1 \\
PL(LN) &48.3±0.2 &43.8±0.1 &25.2±0.1 &57.3±0.2 &40.6±0.3 &30.2±0.2 &32.5±0.0 &44.3±0.0 \\
PL(CLS) &52.0±0.0 &52.6±0.5 &28.2±0.0 &76.6±0.1 &45.1±0.0 &33.8±0.0 &36.9±0.2 &51.3±0.0 \\
PL(Feature) &69.7±7.7 &67.3±14.3 &24.0±0.0 &50.1±0.2 &36.2±0.3 &27.8±0.2 &30.3±0.3 &46.5±0.8 \\
PL(ALL) &72.8±4.8 &68.4±15.1 &24.0±0.2 &50.5±0.3 &36.5±0.4 &27.8±0.2 &30.3±0.2 &46.9±0.7 \\
SHOT-IM(LN) &44.5±0.1 &40.9±0.1 &25.2±0.1 &53.4±0.1 &36.7±0.2 &29.5±0.1 &31.5±0.1 &42.0±0.0 \\
SHOT-IM(CLS) &51.6±0.1 &50.2±0.0 &28.2±0.0 &74.6±0.0 &44.9±0.0 &33.8±0.0 &36.7±0.0 &50.7±0.1 \\
SHOT-IM(Feature) &40.7±0.1 &37.1±0.4 &23.4±0.1 &46.1±0.3 &31.2±0.3 &26.5±0.2 &29.1±0.0 &38.4±0.0 \\
SHOT-IM(ALL) &40.7±0.1 &37.2±0.4 &23.4±0.1 &46.0±0.1 &31.3±0.2 &26.5±0.2 &29.1±0.1 &38.4±0.0 \\
CFA(LN) &42.4±0.0 &38.9±0.1 &23.9±0.1 &51.8±0.1 &33.7±0.1 &27.7±0.0 &30.1±0.1 &40.2±0.0 \\
CFA(CLS) &49.9±0.0 &50.1±0.1 &27.7±0.0 &73.7±0.1 &43.0±0.1 &32.3±0.1 &34.9±0.2 &49.2±0.0 \\
CFA(Feature) &38.9±0.0 &34.7±0.1 &22.6±0.1 &43.2±0.1 &29.2±0.0 &25.5±0.0 &28.4±0.0 &36.6±0.0 \\
CFA(ALL) &38.9±0.0 &34.7±0.1 &22.6±0.1 &43.2±0.1 &29.2±0.0 &25.5±0.0 &28.4±0.0 &36.6±0.0 \\
\bottomrule
\end{tabular}
\caption{Detailed experiment results of Table \ref{table_modulation_study} (Top-1 error rate) based on ViT-L16.}
\label{table_modulation_study_detail_vitL16}
\end{table*}

\begin{table*}[h]
\captionsetup{width=.90\linewidth}
\centering
\begin{tabular}{lrrrrrrrrr}\toprule
Method &Gaussian &Shot &Impulse &Defocus &Glass &Motion &Zoom &Snow \\
\midrule
C10-C(Source) &23.6±0.0 &21.9±0.0 &28.7±0.0 &5.2±0.0 &27.9±0.0 &9.1±0.0 &4.4±0.0 &5.5±0.0 \\
C10-C(Tent) &19.0±0.5 &16.8±0.1 &36.4±1.7 &4.2±0.0 &16.8±0.2 &6.8±0.1 &3.5±0.1 &5.1±0.1 \\
C10-C(PL) &20.8±0.1 &18.6±0.2 &34.4±2.8 &4.5±0.1 &19.5±0.5 &7.6±0.2 &3.9±0.1 &5.3±0.1 \\
C10-C(TFA) &16.0±0.1 &14.4±0.2 &14.7±0.2 &4.2±0.1 &15.5±0.3 &6.6±0.1 &3.5±0.1 &5.0±0.0 \\
C10-C(T3A) &22.2±0.2 &20.5±0.1 &26.0±0.1 &5.3±0.0 &25.4±0.1 &8.8±0.0 &4.4±0.0 &5.6±0.0 \\
C10-C(SHOT-IM) &16.8±0.2 &15.0±0.2 &14.9±0.2 &4.2±0.0 &15.4±0.2 &6.7±0.1 &3.5±0.1 &4.9±0.0 \\
C10-C(CFA-F) &15.6±0.1 &13.9±0.1 &14.2±0.2 &4.4±0.0 &15.1±0.1 &6.7±0.1 &3.5±0.0 &4.9±0.1 \\
C10-C(CFA-C) &15.7±0.1 &14.0±0.2 &14.5±0.3 &4.2±0.1 &14.5±0.2 &6.2±0.1 &3.4±0.1 &4.8±0.0 \\
C10-C(CFA) &15.5±0.0 &13.8±0.2 &14.1±0.3 &4.2±0.1 &14.5±0.2 &6.3±0.2 &3.4±0.1 &4.7±0.0 \\
C100-C(Source) &55.0±0.0 &52.9±0.0 &57.8±0.0 &18.0±0.0 &60.5±0.0 &23.6±0.0 &16.0±0.0 &22.3±0.0 \\
C100-C(Tent) &42.7±0.3 &40.0±0.9 &47.8±8.5 &15.7±0.1 &44.9±0.9 &19.7±0.1 &14.3±0.1 &20.4±0.1 \\
C100-C(PL) &45.9±0.4 &43.4±1.2 &57.8±4.0 &16.3±0.3 &50.8±0.9 &20.8±0.4 &14.9±0.2 &21.2±0.1 \\
C100-C(TFA) &51.3±0.3 &49.4±0.5 &49.5±0.8 &18.1±0.2 &51.8±0.5 &23.0±0.5 &16.3±0.2 &22.9±0.2 \\
C100-C(T3A) &53.6±0.2 &51.6±0.1 &56.4±0.2 &17.8±0.0 &56.8±0.1 &23.4±0.1 &15.9±0.0 &21.7±0.1 \\
C100-C(SHOT-IM) &39.4±0.3 &37.1±0.2 &38.8±0.1 &15.7±0.2 &38.4±0.1 &19.5±0.1 &14.4±0.1 &20.1±0.0 \\
C100-C(CFA-F) &38.7±0.2 &36.2±0.3 &37.3±0.3 &15.9±0.1 &37.4±0.2 &19.5±0.1 &14.1±0.0 &19.9±0.0 \\
C100-C(CFA-C) &39.3±0.4 &37.0±0.2 &38.3±0.3 &15.6±0.1 &38.6±0.1 &19.5±0.1 &14.2±0.1 &20.0±0.0 \\
C100-C(CFA) &38.0±0.3 &35.9±0.2 &36.8±0.3 &15.5±0.1 &36.9±0.2 &19.0±0.2 &14.0±0.1 &19.6±0.1 \\
IN-C(Source) &77.7±0.0 &75.1±0.0 &77.0±0.0 &66.9±0.0 &69.1±0.0 &58.5±0.0 &62.8±0.0 &60.9±0.0 \\
IN-C(Tent) &66.3±8.8 &77.8±1.6 &59.3±0.2 &50.9±0.2 &49.8±0.2 &46.7±0.0 &50.9±0.2 &75.7±2.0 \\
IN-C(PL) &75.5±8.3 &71.7±3.7 &73.7±8.9 &53.2±0.1 &52.4±0.1 &48.6±0.3 &52.4±0.4 &75.3±1.8 \\
IN-C(TFA) &69.7±0.1 &68.0±0.2 &69.8±0.2 &63.2±0.2 &62.6±0.1 &57.7±0.1 &59.2±0.2 &57.0±0.2 \\
IN-C(T3A) &77.8±0.1 &74.9±0.1 &76.8±0.1 &65.5±0.1 &68.3±0.0 &57.7±0.0 &61.7±0.0 &60.1±0.0 \\
IN-C(SHOT-IM) &58.6±0.1 &56.4±0.1 &57.6±0.1 &49.7±0.1 &48.7±0.2 &46.0±0.1 &46.4±0.1 &47.0±0.1 \\
IN-C(CFA-F) &58.8±0.2 &56.7±0.1 &57.7±0.1 &51.6±0.1 &51.5±0.1 &47.9±0.1 &47.1±0.1 &47.0±0.1 \\
IN-C(CFA-C) &58.7±0.1 &56.1±0.1 &57.5±0.1 &50.1±0.1 &49.0±0.2 &45.6±0.1 &46.3±0.2 &46.2±0.1 \\
IN-C(CFA) &56.5±0.1 &54.2±0.1 &55.4±0.1 &48.3±0.0 &47.1±0.0 &44.3±0.0 &44.4±0.2 &44.9±0.1 \\
\midrule
Method &Frost &Fog &Brightness &Contrast &Elastic &Pixelate &Jpeg &Average \\
\midrule
C10-C(Source) &8.4±0.0 &14.2±0.0 &2.4±0.0 &15.2±0.0 &13.9±0.0 &25.5±0.0 &13.3±0.0 &14.6±0.0 \\
C10-C(Tent) &6.0±0.1 &8.8±0.2 &2.3±0.0 &6.5±0.2 &10.3±0.2 &9.1±0.4 &12.2±0.1 &10.9±0.2 \\
C10-C(PL) &6.9±0.2 &10.5±0.4 &2.4±0.0 &7.9±0.4 &11.6±0.4 &11.9±1.2 &12.4±0.3 &11.9±0.0 \\
C10-C(TFA) &5.7±0.0 &8.0±0.0 &2.2±0.0 &6.1±0.1 &10.1±0.1 &7.7±0.1 &12.2±0.1 &8.8±0.0 \\
C10-C(T3A) &8.0±0.0 &13.5±0.0 &2.5±0.0 &14.1±0.0 &13.4±0.0 &22.9±0.2 &13.5±0.0 &13.7±0.0 \\
C10-C(SHOT-IM) &5.8±0.1 &8.0±0.2 &2.3±0.0 &6.1±0.1 &10.0±0.1 &8.1±0.0 &11.8±0.1 &8.9±0.0 \\
C10-C(CFA-F) &5.6±0.1 &8.1±0.1 &2.2±0.0 &6.3±0.1 &9.6±0.0 &7.7±0.1 &12.6±0.1 &8.7±0.0 \\
C10-C(CFA-C) &5.5±0.0 &7.6±0.1 &2.2±0.1 &5.7±0.1 &9.3±0.0 &7.4±0.1 &12.1±0.2 &8.5±0.0 \\
C10-C(CFA) &5.5±0.1 &7.6±0.0 &2.1±0.0 &5.8±0.1 &9.2±0.1 &7.4±0.2 &12.1±0.1 &8.4±0.0 \\
C100-C(Source) &27.5±0.0 &34.2±0.0 &11.9±0.0 &35.3±0.0 &34.8±0.0 &43.3±0.0 &33.7±0.0 &35.1±0.0 \\
C100-C(Tent) &21.5±0.4 &28.0±0.2 &11.2±0.1 &22.1±0.4 &27.9±0.2 &22.9±0.6 &31.5±0.3 &27.4±0.5 \\
C100-C(PL) &23.6±0.4 &29.8±0.0 &11.6±0.1 &25.6±0.6 &30.6±0.3 &26.8±1.3 &32.3±0.4 &30.1±0.5 \\
C100-C(TFA) &24.1±0.2 &34.0±0.2 &12.9±0.2 &27.5±1.0 &34.3±0.8 &31.7±0.9 &35.7±0.1 &32.2±0.2 \\
C100-C(T3A) &26.3±0.1 &33.0±0.1 &11.9±0.0 &33.7±0.1 &33.4±0.1 &41.2±0.2 &33.3±0.1 &34.0±0.0 \\
C100-C(SHOT-IM) &20.8±0.2 &26.8±0.2 &11.2±0.1 &21.5±0.3 &27.2±0.1 &22.3±0.2 &30.5±0.2 &25.6±0.0 \\
C100-C(CFA-F) &20.2±0.1 &25.9±0.1 &11.2±0.1 &21.9±0.1 &26.1±0.2 &21.5±0.2 &32.1±0.0 &25.2±0.0 \\
C100-C(CFA-C) &20.2±0.2 &26.2±0.2 &11.1±0.1 &20.6±0.4 &26.9±0.1 &21.6±0.1 &31.0±0.2 &25.3±0.1 \\
C100-C(CFA) &19.7±0.1 &25.3±0.2 &10.8±0.1 &20.4±0.3 &25.8±0.2 &20.9±0.2 &31.2±0.3 &24.6±0.1 \\
IN-C(Source) &57.6±0.0 &62.9±0.0 &31.6±0.0 &88.9±0.0 &51.9±0.0 &45.3±0.0 &42.9±0.0 &61.9±0.0 \\
IN-C(Tent) &48.2±0.4 &44.3±0.3 &26.1±0.1 &58.5±0.3 &37.6±0.3 &32.7±0.1 &34.7±0.1 &50.6±0.5 \\
IN-C(PL) &51.4±0.4 &79.5±3.3 &26.8±0.1 &62.8±0.2 &41.1±0.2 &35.1±0.2 &36.2±0.1 &55.7±1.4 \\
IN-C(TFA) &55.1±0.1 &58.3±0.1 &34.7±0.1 &74.8±0.2 &45.9±0.0 &45.5±0.1 &45.1±0.0 &57.8±0.1 \\
IN-C(T3A) &56.8±0.0 &62.0±0.0 &30.7±0.1 &89.4±0.1 &49.9±0.0 &44.7±0.0 &41.7±0.0 &61.2±0.0 \\
IN-C(SHOT-IM) &45.9±0.1 &43.2±0.2 &26.1±0.0 &56.9±0.2 &35.9±0.2 &32.4±0.1 &34.3±0.1 &45.7±0.0 \\
IN-C(CFA-F) &47.0±0.1 &44.4±0.1 &27.0±0.0 &58.9±0.1 &35.4±0.0 &33.4±0.0 &35.5±0.0 &46.7±0.0 \\
IN-C(CFA-C) &45.6±0.1 &41.9±0.2 &26.1±0.1 &56.3±0.2 &34.7±0.0 &31.5±0.1 &34.2±0.0 &45.3±0.0 \\
IN-C(CFA) &44.8±0.1 &41.2±0.1 &25.6±0.1 &54.4±0.2 &33.2±0.1 &30.5±0.0 &33.5±0.1 &43.9±0.0 \\
\bottomrule
\end{tabular}
\caption{Detailed experiment results of Table \ref{table_method_comparison} (Top-1 error rate) for each methods on CIFAR-10-C, CIFAR-100-C and ImageNet-C with severity level=5 by ViT-B16 model.}
\label{table_method_comparison_detail}
\end{table*}

\begin{table*}[h]\centering
\begin{tabular}{lrrrrrrrrr}\toprule
Method &Gaussian &Shot &Impulse &Defocus &Glass &Motion &Zoom &Snow \\
\midrule
ResNet50 &97.8±0.0 &97.1±0.0 &98.1±0.0 &82.1±0.0 &90.2±0.0 &85.2±0.0 &77.5±0.0 &83.1±0.0 \\
ResNet101 &96.5±0.0 &95.7±0.0 &96.5±0.0 &78.1±0.0 &86.8±0.0 &80.8±0.0 &73.5±0.0 &79.0±0.0 \\
ViT-B16 &77.7±0.0 &75.1±0.0 &77.0±0.0 &66.9±0.0 &69.1±0.0 &58.5±0.0 &62.8±0.0 &60.9±0.0 \\
ViT-L16 &60.5±0.0 &59.2±0.0 &59.1±0.0 &61.3±0.0 &62.6±0.0 &51.0±0.0 &54.2±0.0 &53.4±0.0 \\
DeiT-S16 &69.6±0.0 &67.6±0.0 &68.1±0.0 &72.4±0.0 &81.8±0.0 &65.2±0.0 &70.9±0.0 &51.4±0.0 \\
DeiT-B16 &58.0±0.0 &57.6±0.0 &57.0±0.0 &66.7±0.0 &76.8±0.0 &60.8±0.0 &65.2±0.0 &45.0±0.0 \\
MLP-Mixer-B16 &85.0±0.0 &84.6±0.0 &87.1±0.0 &82.3±0.0 &88.3±0.0 &72.7±0.0 &77.0±0.0 &67.6±0.0 \\
MLP-Mixer-L16 &82.8±0.0 &84.9±0.0 &85.8±0.0 &86.0±0.0 &90.2±0.0 &80.3±0.0 &80.2±0.0 &79.0±0.0 \\
ViT-B16 (AugReg) &53.1±0.0 &52.4±0.0 &53.1±0.0 &57.3±0.0 &65.8±0.0 &49.6±0.0 &55.3±0.0 &43.1±0.0 \\
ViT-L16 (AugReg) &37.9±0.0 &38.6±0.0 &37.7±0.0 &47.3±0.0 &54.9±0.0 &39.4±0.0 &44.9±0.0 &33.8±0.0 \\
BeiT-B16 &52.8±0.0 &49.9±0.0 &50.5±0.0 &57.7±0.0 &65.9±0.0 &50.4±0.0 &57.0±0.0 &41.8±0.0 \\
BeiT-L16 &36.5±0.0 &35.1±0.0 &35.0±0.0 &42.5±0.0 &53.0±0.0 &36.7±0.0 &43.1±0.0 &29.8±0.0 \\
\midrule
Method &Frost &Fog &Brightness &Contrast &Elastic &Pixelate &Jpeg &Average \\
\midrule
ResNet50 &76.7±0.0 &75.6±0.0 &41.1±0.0 &94.6±0.0 &83.1±0.0 &79.4±0.0 &68.3±0.0 &82.0±0.0 \\
ResNet101 &73.3±0.0 &71.9±0.0 &38.6±0.0 &92.8±0.0 &75.7±0.0 &65.0±0.0 &57.6±0.0 &77.4±0.0 \\
ViT-B16 &57.6±0.0 &62.9±0.0 &31.6±0.0 &88.9±0.0 &51.9±0.0 &45.3±0.0 &42.9±0.0 &61.9±0.0 \\
ViT-L16 &52.6±0.0 &57.6±0.0 &28.3±0.0 &83.8±0.0 &45.3±0.0 &34.1±0.0 &37.5±0.0 &53.4±0.0 \\
DeiT-S16 &49.1±0.0 &41.6±0.0 &27.4±0.0 &52.5±0.0 &66.9±0.0 &65.2±0.0 &48.6±0.0 &59.9±0.0 \\
DeiT-B16 &42.6±0.0 &37.1±0.0 &24.0±0.0 &45.2±0.0 &64.1±0.0 &51.1±0.0 &41.5±0.0 &52.9±0.0 \\
MLP-Mixer-B16 &59.1±0.0 &62.0±0.0 &35.8±0.0 &88.3±0.0 &75.4±0.0 &71.6±0.0 &62.4±0.0 &73.3±0.0 \\
MLP-Mixer-L16 &65.0±0.0 &63.0±0.0 &43.8±0.0 &87.6±0.0 &75.3±0.0 &82.7±0.0 &69.1±0.0 &77.1±0.0 \\
ViT-B16 (AugReg) &47.4±0.0 &43.5±0.0 &23.9±0.0 &68.2±0.0 &53.3±0.0 &34.5±0.0 &34.0±0.0 &49.0±0.0 \\
ViT-L16 (AugReg) &37.6±0.0 &37.5±0.0 &19.8±0.0 &60.1±0.0 &43.8±0.0 &25.7±0.0 &27.3±0.0 &39.1±0.0 \\
BeiT-B16 &48.5±0.0 &52.5±0.0 &23.3±0.0 &47.8±0.0 &57.7±0.0 &36.1±0.0 &33.1±0.0 &48.3±0.0 \\
BeiT-L16 &37.0±0.0 &37.6±0.0 &18.5±0.0 &33.5±0.0 &46.8±0.0 &27.0±0.0 &26.4±0.0 &35.9±0.0 \\
\bottomrule
\end{tabular}
\caption{Detailed experiment results of Table \ref{table_model_agnosticity} (Top-1 error rate) by Source.}
\label{table_model_agnosticity_detail_1}
\end{table*}

\begin{table*}[h]\centering
\begin{tabular}{lrrrrrrrrr}\toprule
Method &Gaussian &Shot &Impulse &Defocus &Glass &Motion &Zoom &Snow \\
\midrule
ResNet50 &74.2±0.1 &72.2±0.1 &73.4±0.1 &74.4±0.1 &74.7±0.3 &61.5±0.1 &52.0±0.1 &54.7±0.1 \\
ResNet101 &71.2±0.1 &68.7±0.1 &70.8±0.0 &70.8±0.2 &70.5±0.2 &57.9±0.1 &48.8±0.0 &51.7±0.1 \\
ViT-B16 &58.6±0.1 &56.4±0.1 &57.7±0.1 &49.8±0.2 &48.7±0.2 &45.9±0.1 &46.4±0.1 &47.0±0.1 \\
ViT-L16 &48.8±0.0 &47.5±0.1 &47.9±0.1 &45.9±0.0 &47.6±0.1 &41.7±0.1 &43.8±0.1 &44.5±0.1 \\
DeiT-S16 &54.4±0.2 &52.4±0.1 &53.3±0.1 &57.3±0.1 &56.1±0.1 &48.9±0.1 &51.4±0.0 &42.6±0.0 \\
DeiT-B16 &43.7±0.1 &42.6±0.1 &42.5±0.1 &52.0±0.1 &49.9±0.0 &43.6±0.2 &45.8±0.2 &36.1±0.0 \\
MLP-Mixer-B16 &66.2±0.1 &65.4±0.2 &65.7±0.3 &65.4±0.1 &65.4±0.5 &55.8±0.2 &55.7±0.1 &51.0±0.3 \\
MLP-Mixer-L16 &65.8±0.2 &65.6±0.1 &66.6±0.2 &72.7±0.4 &73.8±0.4 &62.7±0.0 &66.0±0.2 &60.9±0.0 \\
ViT-B16 (AugReg) &43.4±0.0 &42.3±0.0 &42.4±0.1 &44.1±0.1 &48.5±0.0 &40.9±0.1 &43.9±0.1 &35.3±0.2 \\
ViT-L16 (AugReg) &34.0±0.0 &32.9±0.1 &33.4±0.1 &38.2±0.1 &43.3±0.2 &34.3±0.1 &38.2±0.0 &30.6±0.0 \\
BeiT-B16 &43.3±0.1 &40.9±0.0 &41.9±0.1 &44.9±0.1 &46.2±0.1 &40.0±0.1 &43.8±0.1 &35.1±0.0 \\
BeiT-L16 &30.6±0.0 &29.4±0.0 &29.7±0.0 &34.1±0.1 &36.9±0.2 &30.3±0.1 &34.0±0.1 &25.3±0.1 \\
\midrule
Method &Frost &Fog &Brightness &Contrast &Elastic &Pixelate &Jpeg &Average \\
\midrule
ResNet50 &59.7±0.1 &43.6±0.1 &32.8±0.1 &70.5±0.2 &46.2±0.1 &42.4±0.1 &49.1±0.0 &58.8±0.0 \\
ResNet101 &56.6±0.1 &42.0±0.1 &31.2±0.0 &67.2±0.1 &42.5±0.1 &39.8±0.1 &45.9±0.1 &55.7±0.0 \\
ViT-B16 &45.9±0.1 &43.2±0.2 &26.1±0.0 &56.9±0.2 &35.9±0.2 &32.4±0.1 &34.4±0.1 &45.7±0.0 \\
ViT-L16 &44.5±0.1 &40.9±0.1 &25.2±0.1 &53.4±0.1 &36.7±0.1 &29.5±0.1 &31.5±0.1 &42.0±0.0 \\
DeiT-S16 &43.4±0.1 &35.3±0.1 &26.1±0.0 &41.3±0.1 &47.1±0.1 &42.6±0.1 &39.7±0.1 &46.1±0.0 \\
DeiT-B16 &39.4±0.1 &31.1±0.0 &23.1±0.0 &35.5±0.1 &43.0±0.2 &35.6±0.1 &34.8±0.1 &39.9±0.0 \\
MLP-Mixer-B16 &51.4±0.3 &47.5±0.6 &31.2±0.1 &60.7±0.3 &49.9±0.1 &46.6±0.2 &48.3±0.2 &55.1±0.1 \\
MLP-Mixer-L16 &63.0±0.7 &57.7±0.1 &39.9±0.2 &70.5±0.2 &59.3±0.6 &55.1±0.5 &56.6±0.2 &62.4±0.1 \\
ViT-B16 (AugReg) &37.4±0.1 &33.8±0.1 &22.4±0.0 &42.0±0.2 &39.2±0.0 &29.8±0.1 &31.5±0.1 &38.4±0.0 \\
ViT-L16 (AugReg) &33.3±0.1 &32.5±0.1 &19.3±0.0 &41.5±0.0 &37.0±0.1 &23.9±0.0 &26.2±0.2 &33.3±0.0 \\
BeiT-B16 &39.0±0.1 &35.4±0.1 &21.8±0.1 &33.0±0.2 &39.0±0.1 &30.2±0.1 &29.4±0.0 &37.6±0.0 \\
BeiT-L16 &30.3±0.0 &26.7±0.1 &16.7±0.0 &23.6±0.1 &30.8±0.1 &21.5±0.1 &22.9±0.0 &28.2±0.0 \\
\bottomrule
\end{tabular}
\caption{Detailed experiment results of Table \ref{table_model_agnosticity} (Top-1 error rate) by SHOT-IM.}
\label{table_model_agnosticity_detail_2}
\end{table*}

\begin{table*}[h]\centering
\begin{tabular}{lrrrrrrrrr}\toprule
Method &Gaussian &Shot &Impulse &Defocus &Glass &Motion &Zoom &Snow \\
\midrule
ResNet50 &74.6±0.0 &73.7±0.1 &73.7±0.2 &74.6±0.1 &75.0±0.1 &61.4±0.1 &52.1±0.0 &54.3±0.1 \\
ResNet101 &70.8±0.1 &68.5±0.1 &70.5±0.1 &70.6±0.2 &70.2±0.1 &57.5±0.2 &48.7±0.0 &51.1±0.0 \\
ViT-B16 &56.5±0.1 &54.1±0.1 &55.4±0.1 &48.4±0.0 &47.1±0.0 &44.3±0.0 &44.4±0.2 &44.9±0.1 \\
ViT-L16 &47.3±0.0 &45.9±0.1 &46.4±0.1 &44.6±0.1 &45.8±0.0 &40.3±0.1 &41.9±0.0 &42.2±0.1 \\
DeiT-S16 &54.5±0.1 &52.4±0.1 &53.5±0.1 &58.0±0.1 &56.8±0.2 &49.5±0.1 &51.4±0.1 &41.9±0.1 \\
DeiT-B16 &44.5±0.1 &42.9±0.1 &43.3±0.0 &54.2±0.1 &53.3±0.1 &44.3±0.1 &45.4±0.2 &34.1±0.1 \\
MLP-Mixer-B16 &63.9±0.1 &62.5±0.1 &63.3±0.0 &64.5±0.1 &66.4±0.2 &54.3±0.2 &54.1±0.2 &46.3±0.1 \\
MLP-Mixer-L16 &61.1±0.1 &60.1±0.3 &61.8±0.1 &70.6±0.1 &69.7±0.1 &57.9±0.2 &58.7±0.2 &52.0±0.2 \\
ViT-B16 (AugReg) &43.1±0.0 &42.0±0.1 &41.9±0.1 &45.6±0.0 &51.1±0.2 &40.1±0.0 &43.3±0.1 &33.6±0.2 \\
ViT-L16 (AugReg) &33.6±0.1 &32.8±0.1 &33.3±0.1 &38.1±0.1 &42.5±0.2 &34.2±0.2 &36.5±0.1 &28.6±0.0 \\
BeiT-B16 &41.9±0.1 &39.5±0.1 &40.5±0.1 &43.2±0.0 &44.2±0.1 &37.7±0.1 &41.2±0.1 &32.7±0.0 \\
BeiT-L16 &29.5±0.1 &28.0±0.1 &28.5±0.0 &32.0±0.0 &33.7±0.1 &27.7±0.1 &30.2±0.0 &23.9±0.0 \\
\midrule
Method &Frost &Fog &Brightness &Contrast &Elastic &Pixelate &Jpeg &Average \\
\midrule
ResNet50 &59.5±0.1 &43.8±0.0 &33.2±0.1 &68.6±0.2 &46.2±0.1 &42.4±0.2 &49.2±0.1 &58.8±0.0 \\
ResNet101 &56.0±0.1 &41.9±0.1 &31.4±0.1 &64.4±0.2 &42.2±0.2 &39.7±0.1 &45.7±0.1 &55.3±0.1 \\
ViT-B16 &44.8±0.1 &41.1±0.1 &25.6±0.1 &54.4±0.3 &33.3±0.1 &30.6±0.0 &33.5±0.1 &43.9±0.0 \\
ViT-L16 &42.4±0.0 &38.9±0.1 &23.9±0.1 &51.8±0.1 &33.7±0.1 &27.7±0.0 &30.1±0.1 &40.2±0.0 \\
DeiT-S16 &42.9±0.1 &38.5±0.2 &25.0±0.0 &41.3±0.1 &46.0±0.1 &40.8±0.1 &38.2±0.0 &46.0±0.0 \\
DeiT-B16 &37.4±0.0 &35.2±0.2 &21.7±0.0 &34.2±0.0 &41.0±0.1 &34.1±0.0 &32.8±0.1 &39.9±0.0 \\
MLP-Mixer-B16 &46.8±0.1 &42.5±0.3 &29.3±0.1 &53.3±0.1 &48.7±0.5 &44.8±0.3 &45.7±0.1 &52.4±0.1 \\
MLP-Mixer-L16 &55.2±0.3 &48.6±0.0 &33.9±0.1 &60.6±0.6 &52.3±0.3 &49.2±0.1 &52.5±0.1 &56.3±0.0 \\
ViT-B16 (AugReg) &35.9±0.2 &32.3±0.0 &21.0±0.0 &41.2±0.1 &35.7±0.1 &28.3±0.0 &29.8±0.1 &37.6±0.0 \\
ViT-L16 (AugReg) &31.0±0.1 &30.7±0.0 &18.1±0.0 &41.4±0.3 &32.8±0.1 &22.2±0.1 &25.4±0.1 &32.1±0.0 \\
BeiT-B16 &37.1±0.1 &33.4±0.1 &19.7±0.1 &31.0±0.1 &35.1±0.1 &27.3±0.1 &26.9±0.0 &35.4±0.0 \\
BeiT-L16 &28.8±0.1 &24.4±0.1 &15.7±0.0 &22.1±0.0 &26.4±0.1 &19.4±0.0 &20.7±0.0 &26.0±0.0 \\
\bottomrule
\end{tabular}
\caption{Detailed experiment results of Table \ref{table_model_agnosticity} (Top-1 error rate) by CFA.}
\label{table_model_agnosticity_detail_3}
\end{table*}

\begin{table*}[h]\centering
\begin{tabular}{lrrrrrrrrr}\toprule
Method &Gaussian &Shot &Impulse &Defocus &Glass &Motion &Zoom &Snow \\
\midrule
Source (s=1) &16.8±0.0 &16.7±0.0 &16.8±0.0 &17.4±0.0 &17.5±0.0 &15.3±0.0 &19.9±0.0 &17.9±0.0 \\
Source (s=2) &17.9±0.0 &18.2±0.0 &18.4±0.0 &19.9±0.0 &20.7±0.0 &17.1±0.0 &24.3±0.0 &23.5±0.0 \\
Source (s=3) &20.3±0.0 &20.5±0.0 &19.9±0.0 &26.1±0.0 &35.3±0.0 &21.6±0.0 &29.6±0.0 &23.0±0.0 \\
Source (s=4) &25.5±0.0 &27.0±0.0 &25.5±0.0 &33.6±0.0 &40.9±0.0 &29.5±0.0 &34.7±0.0 &27.1±0.0 \\
Source (s=5) &36.5±0.0 &35.1±0.0 &35.0±0.0 &42.5±0.0 &53.0±0.0 &36.7±0.0 &43.1±0.0 &29.8±0.0 \\
Tent (s=1) &15.3±0.1 &15.4±0.0 &15.7±0.0 &15.4±0.0 &15.7±0.0 &14.5±0.1 &18.1±0.0 &16.4±0.0 \\
Tent (s=2) &16.4±0.0 &16.6±0.0 &17.1±0.0 &17.0±0.0 &17.8±0.1 &16.1±0.0 &21.1±0.1 &20.6±0.0 \\
Tent (s=3) &18.6±0.0 &18.7±0.0 &18.6±0.0 &21.2±0.0 &25.4±0.1 &19.5±0.1 &24.9±0.0 &20.4±0.0 \\
Tent (s=4) &22.5±0.1 &23.5±0.0 &22.7±0.0 &26.6±0.0 &28.8±0.1 &24.9±0.1 &28.0±0.0 &23.4±0.1 \\
Tent (s=5) &29.9±0.0 &28.6±0.1 &29.2±0.1 &33.7±0.1 &36.5±0.2 &29.6±0.1 &33.8±0.1 &24.8±0.1 \\
SHOT-IM (s=1) &15.8±0.0 &15.9±0.0 &16.1±0.0 &15.7±0.0 &16.2±0.0 &14.7±0.0 &18.6±0.0 &16.7±0.0 \\
SHOT-IM (s=2) &16.6±0.0 &17.0±0.0 &17.4±0.0 &17.5±0.0 &18.6±0.0 &16.3±0.0 &21.9±0.1 &21.1±0.0 \\
SHOT-IM (s=3) &18.9±0.0 &19.1±0.0 &18.9±0.0 &21.7±0.0 &26.8±0.1 &20.0±0.0 &25.7±0.0 &21.0±0.1 \\
SHOT-IM (s=4) &23.2±0.1 &24.2±0.1 &23.3±0.0 &27.2±0.1 &30.1±0.1 &25.6±0.1 &28.8±0.0 &24.0±0.0 \\
SHOT-IM (s=5) &30.6±0.0 &29.4±0.0 &29.7±0.0 &34.1±0.1 &36.9±0.2 &30.3±0.1 &34.0±0.1 &25.3±0.1 \\
CFA (s=1) &15.2±0.1 &15.2±0.0 &15.7±0.0 &15.3±0.0 &15.4±0.0 &14.6±0.1 &17.6±0.1 &16.2±0.0 \\
CFA (s=2) &16.3±0.1 &16.5±0.0 &17.0±0.1 &16.8±0.1 &17.2±0.1 &15.9±0.1 &20.1±0.0 &20.0±0.1 \\
CFA (s=3) &18.4±0.0 &18.6±0.1 &18.4±0.1 &20.5±0.0 &23.6±0.0 &19.0±0.0 &23.1±0.1 &19.7±0.0 \\
CFA (s=4) &22.3±0.1 &23.1±0.1 &22.4±0.1 &25.6±0.1 &26.7±0.1 &23.5±0.1 &25.8±0.1 &22.7±0.1 \\
CFA (s=5) &29.5±0.1 &28.0±0.1 &28.5±0.0 &32.0±0.0 &33.7±0.1 &27.7±0.1 &30.2±0.0 &23.9±0.0 \\
\midrule
Method &Frost &Fog &Brightness &Contrast &Elastic &Pixelate &Jpeg &Average \\
\midrule
Source (s=1) &18.1±0.0 &17.8±0.0 &14.4±0.0 &14.7±0.0 &17.5±0.0 &15.7±0.0 &16.2±0.0 &16.8±0.0 \\
Source (s=2) &24.5±0.0 &19.5±0.0 &15.0±0.0 &15.1±0.0 &36.1±0.0 &16.9±0.0 &17.5±0.0 &20.3±0.0 \\
Source (s=3) &30.6±0.0 &22.8±0.0 &15.8±0.0 &16.3±0.0 &18.9±0.0 &17.7±0.0 &18.5±0.0 &22.5±0.0 \\
Source (s=4) &32.1±0.0 &26.9±0.0 &16.7±0.0 &20.9±0.0 &24.6±0.0 &20.5±0.0 &21.7±0.0 &27.2±0.0 \\
Source (s=5) &37.0±0.0 &37.6±0.0 &18.5±0.0 &33.5±0.0 &46.8±0.0 &27.0±0.0 &26.4±0.0 &35.9±0.0 \\
Tent (s=1) &16.3±0.1 &15.9±0.0 &13.4±0.0 &14.0±0.1 &15.8±0.0 &14.2±0.0 &15.0±0.0 &15.4±0.0 \\
Tent (s=2) &21.2±0.0 &16.8±0.0 &13.8±0.0 &14.3±0.0 &29.6±0.0 &14.7±0.0 &15.8±0.1 &17.9±0.0 \\
Tent (s=3) &30.9±6.4 &18.6±0.1 &14.4±0.0 &15.1±0.0 &16.6±0.0 &15.7±0.0 &16.3±0.1 &19.6±0.4 \\
Tent (s=4) &50.2±17.7 &20.9±0.1 &15.0±0.1 &17.4±0.0 &19.5±0.0 &17.8±0.1 &18.5±0.0 &24.0±1.2 \\
Tent (s=5) &75.5±3.4 &26.1±0.0 &16.0±0.1 &23.0±0.0 &75.4±4.0 &20.3±0.1 &21.5±0.0 &33.6±0.1 \\
SHOT-IM (s=1) &16.6±0.0 &16.4±0.1 &13.6±0.1 &14.2±0.0 &16.2±0.0 &14.4±0.0 &15.5±0.0 &15.8±0.0 \\
SHOT-IM (s=2) &21.6±0.0 &17.4±0.0 &14.0±0.1 &14.6±0.0 &30.1±0.1 &15.1±0.1 &16.4±0.1 &18.4±0.0 \\
SHOT-IM (s=3) &25.9±0.1 &19.4±0.0 &14.6±0.0 &15.4±0.0 &17.3±0.0 &16.1±0.0 &17.0±0.0 &19.9±0.0 \\
SHOT-IM (s=4) &27.0±0.1 &21.6±0.0 &15.4±0.1 &18.1±0.1 &20.8±0.0 &18.6±0.1 &19.5±0.1 &23.2±0.0 \\
SHOT-IM (s=5) &30.3±0.0 &26.7±0.1 &16.7±0.0 &23.6±0.1 &30.8±0.1 &21.5±0.1 &22.9±0.0 &28.2±0.0 \\
CFA (s=1) &16.1±0.1 &15.6±0.0 &13.5±0.1 &14.0±0.0 &15.6±0.0 &14.2±0.0 &14.8±0.0 &15.3±0.0 \\
CFA (s=2) &20.4±0.0 &16.3±0.0 &13.8±0.1 &14.4±0.0 &27.7±0.0 &14.4±0.0 &15.5±0.1 &17.5±0.0 \\
CFA (s=3) &24.4±0.1 &17.8±0.0 &14.2±0.0 &14.9±0.0 &16.1±0.0 &15.4±0.0 &16.0±0.0 &18.7±0.0 \\
CFA (s=4) &25.5±0.1 &19.7±0.1 &14.9±0.0 &16.8±0.0 &18.1±0.0 &17.3±0.1 &17.7±0.1 &21.5±0.0 \\
CFA (s=5) &28.8±0.1 &24.4±0.1 &15.7±0.0 &22.1±0.0 &26.4±0.1 &19.4±0.0 &20.7±0.0 &26.0±0.0 \\
\bottomrule
\end{tabular}
\caption{Detailed experiment results of Table \ref{table_beit_result} (Top-1 error rate).}
\label{table_beit_result_detail}
\end{table*}

\begin{table*}[h]\centering
\begin{tabular}{lrrrrrrrrr}\toprule
Method &Gaussian &Shot &Impulse &Defocus &Glass &Motion &Zoom &Snow \\
\midrule
Tent (LR=0.01) &96.1±1.8 &95.6±1.2 &93.4±5.9 &66.6±5.7 &70.6±11.5 &67.8±13.0 &87.6±3.0 &96.5±1.0 \\
Tent (LR=0.0001) &69.1±0.3 &66.8±0.0 &68.2±0.1 &62.1±0.1 &63.0±0.1 &53.8±0.1 &57.6±0.1 &56.2±0.1 \\
Tent (BS=32) &71.7±5.5 &81.9±0.6 &63.3±8.0 &49.8±0.4 &48.3±0.1 &45.7±0.1 &56.2±4.1 &83.2±3.0 \\
Tent (BS=128) &62.5±1.1 &67.4±4.6 &61.2±0.7 &52.5±0.1 &51.6±0.3 &47.7±0.0 &51.2±0.1 &62.9±5.1 \\
Tent (GC=OFF) &97.3±0.3 &92.0±2.6 &94.9±3.9 &53.9±3.5 &59.6±16.1 &50.9±1.3 &85.9±0.8 &93.2±5.2 \\
CFA (LR=0.01) &54.0±0.3 &51.4±0.3 &52.9±0.1 &47.3±0.2 &44.6±0.3 &40.8±0.0 &40.3±0.1 &41.1±0.3 \\
CFA (LR=0.0001) &66.3±0.2 &63.7±0.1 &65.2±0.1 &57.7±0.0 &59.4±0.0 &51.6±0.0 &54.0±0.1 &53.4±0.1 \\
CFA (BS=32) &55.6±0.1 &53.2±0.1 &54.6±0.1 &47.5±0.2 &46.2±0.1 &43.4±0.1 &43.4±0.2 &44.1±0.1 \\
CFA (BS=128) &58.2±0.1 &55.8±0.1 &57.2±0.1 &50.0±0.1 &49.0±0.0 &45.8±0.0 &46.1±0.1 &46.4±0.1 \\
CFA (GC=OFF) &53.9±0.2 &51.5±0.2 &52.8±0.1 &46.6±0.1 &44.6±0.2 &41.2±0.1 &40.9±0.0 &41.6±0.1 \\
CFA ($\lambda$=0.5) &56.5±0.1 &54.2±0.1 &55.5±0.1 &48.7±0.0 &47.4±0.1 &44.6±0.1 &44.6±0.2 &44.9±0.1 \\
CFA ($\lambda$=2.0) &56.8±0.2 &54.5±0.0 &55.9±0.1 &48.6±0.1 &47.5±0.1 &44.5±0.1 &44.7±0.2 &45.1±0.1 \\
\midrule
Method &Frost &Fog &Brightness &Contrast &Elastic &Pixelate &Jpeg &Average \\
\midrule
Tent (LR=0.01) &89.9±2.6 &91.1±2.5 &26.1±0.2 &84.0±3.4 &35.5±3.6 &30.2±0.2 &33.8±0.4 &71.0±1.5 \\
Tent (LR=0.0001) &54.7±0.1 &55.1±0.0 &29.3±0.1 &76.8±0.2 &48.4±0.1 &41.2±0.0 &40.0±0.1 &56.2±0.0 \\
Tent (BS=32) &47.6±0.9 &42.9±0.2 &25.7±0.1 &56.5±0.5 &35.9±0.4 &31.5±0.1 &34.0±0.2 &51.6±0.8 \\
Tent (BS=128) &49.3±0.1 &46.1±0.2 &26.5±0.1 &62.1±0.1 &39.6±0.3 &34.3±0.0 &35.5±0.1 &50.0±0.6 \\
Tent (GC=OFF) &83.5±2.5 &84.9±7.4 &25.5±0.0 &85.2±8.6 &33.4±0.5 &29.8±0.0 &33.1±0.1 &66.9±1.5 \\
CFA (LR=0.01) &42.9±0.2 &36.6±0.0 &25.4±0.1 &46.6±0.3 &30.7±0.1 &28.5±0.0 &32.4±0.1 &41.0±0.1 \\
CFA (LR=0.0001) &51.8±0.0 &52.4±0.1 &28.3±0.0 &73.1±0.1 &44.4±0.0 &39.0±0.1 &38.2±0.1 &53.2±0.0 \\
CFA (BS=32) &44.1±0.1 &39.8±0.0 &25.5±0.1 &52.0±0.0 &32.4±0.0 &29.8±0.1 &33.0±0.1 &43.0±0.0 \\
CFA (BS=128) &45.9±0.1 &43.4±0.1 &26.0±0.0 &57.8±0.1 &35.1±0.1 &32.0±0.1 &34.3±0.1 &45.5±0.0 \\
CFA (GC=OFF) &42.5±0.2 &36.9±0.1 &25.3±0.1 &46.9±0.1 &30.8±0.1 &28.6±0.1 &32.1±0.1 &41.1±0.0 \\
CFA ($\lambda$=0.5) &45.0±0.0 &41.4±0.0 &25.6±0.1 &54.6±0.2 &33.3±0.1 &30.8±0.0 &33.7±0.1 &44.1±0.0 \\
CFA ($\lambda$=2.0) &44.8±0.0 &41.2±0.1 &25.7±0.1 &54.5±0.2 &33.5±0.1 &30.7±0.0 &33.6±0.1 &44.1±0.0 \\
\bottomrule
\end{tabular}
\caption{Detailed experiment results of Figure \ref{fig_hypara} (Top-1 error rate).}
\label{table_hypara_detail}
\end{table*}

\begin{table*}[h]\centering
\begin{tabular}{lrrrrrrrrr}\toprule
Method &Gaussian &Shot &Impulse &Defocus &Glass &Motion &Zoom &Snow \\
\midrule
CFA-F ($K$=1) ($h$) &58.9±0.1 &56.7±0.1 &57.7±0.1 &51.6±0.0 &51.6±0.0 &48.0±0.1 &47.1±0.1 &47.0±0.1 \\
CFA-F ($K$=1) ($f$) &58.6±0.1 &56.5±0.0 &57.5±0.1 &51.5±0.1 &51.8±0.2 &47.7±0.0 &46.8±0.1 &46.7±0.1 \\
CFA-F ($K$=3) ($h$) &58.8±0.2 &56.7±0.1 &57.7±0.1 &51.6±0.1 &51.5±0.1 &47.9±0.1 &47.1±0.1 &47.0±0.1 \\
CFA-F ($K$=3) ($f$) &59.4±0.3 &57.1±0.2 &58.4±0.3 &51.5±0.1 &52.5±0.3 &48.4±0.0 &47.8±0.2 &47.9±0.1 \\
CFA-F ($K$=5) ($h$) &58.8±0.2 &56.7±0.1 &57.6±0.1 &51.6±0.1 &51.5±0.0 &47.8±0.0 &47.0±0.1 &47.1±0.1 \\
CFA-F ($K$=5) ($f$) &66.1±0.3 &64.0±0.2 &65.5±0.2 &59.1±0.2 &62.1±1.4 &53.5±0.2 &55.9±0.5 &56.4±0.5 \\
CFA-C ($h$) &58.7±0.1 &56.1±0.1 &57.5±0.1 &50.1±0.1 &49.0±0.2 &45.6±0.1 &46.3±0.2 &46.2±0.1 \\
CFA-C ($f$) &60.7±1.1 &58.8±0.2 &59.1±1.0 &52.0±0.2 &50.9±0.0 &46.8±0.1 &47.0±0.3 &47.0±0.0 \\
CFA ($K$=1) ($h$) &56.6±0.1 &54.3±0.2 &55.6±0.1 &48.5±0.0 &47.2±0.1 &44.4±0.0 &44.5±0.2 &45.0±0.1 \\
CFA ($K$=1) ($f$) &57.3±0.0 &54.9±0.0 &56.2±0.1 &50.4±0.1 &49.7±0.0 &45.8±0.0 &45.5±0.1 &45.7±0.1 \\
CFA ($K$=3) ($h$) &56.5±0.1 &54.2±0.1 &55.4±0.1 &48.3±0.0 &47.1±0.0 &44.3±0.0 &44.4±0.2 &44.9±0.1 \\
CFA ($K$=3) ($f$) &57.2±0.1 &54.9±0.0 &56.1±0.1 &48.5±0.1 &48.1±0.1 &44.9±0.1 &45.2±0.1 &45.6±0.0 \\
CFA ($K$=5) ($h$) &56.4±0.1 &54.2±0.1 &55.5±0.1 &48.4±0.1 &47.2±0.1 &44.4±0.0 &44.4±0.2 &44.9±0.1 \\
CFA ($K$=5) ($f$) &64.6±0.3 &62.0±0.1 &63.5±0.1 &55.7±0.3 &58.7±0.8 &51.3±0.2 &53.4±0.2 &54.0±0.3 \\
\midrule
Method &Frost &Fog &Brightness &Contrast &Elastic &Pixelate &Jpeg &Average \\
\midrule
CFA-F ($K$=1) ($h$) &47.0±0.1 &44.4±0.1 &27.0±0.0 &58.9±0.1 &35.4±0.0 &33.4±0.0 &35.6±0.0 &46.7±0.0 \\
CFA-F ($K$=1) ($f$) &46.8±0.0 &44.2±0.1 &26.8±0.0 &61.7±0.4 &35.2±0.0 &33.2±0.1 &35.3±0.0 &46.7±0.0 \\
CFA-F ($K$=3) ($h$) &47.0±0.1 &44.4±0.1 &27.0±0.0 &58.9±0.1 &35.4±0.0 &33.4±0.0 &35.5±0.0 &46.7±0.0 \\
CFA-F ($K$=3) ($f$) &47.4±0.1 &45.6±0.1 &27.1±0.1 &59.9±0.3 &36.6±0.1 &34.0±0.3 &35.6±0.1 &47.3±0.0 \\
CFA-F ($K$=5) ($h$) &47.0±0.1 &44.4±0.0 &27.0±0.0 &58.8±0.2 &35.4±0.0 &33.4±0.0 &35.5±0.0 &46.6±0.0 \\
CFA-F ($K$=5) ($f$) &53.4±0.5 &53.5±0.3 &29.4±0.4 &72.6±0.3 &45.8±0.1 &40.7±0.4 &39.7±0.3 &54.5±0.1 \\
CFA-C ($h$) &45.6±0.1 &41.9±0.2 &26.1±0.1 &56.3±0.2 &34.7±0.0 &31.5±0.1 &34.2±0.0 &45.3±0.0 \\
CFA-C ($f$) &46.7±0.1 &43.0±0.2 &26.5±0.1 &65.9±0.1 &35.4±0.0 &32.0±0.1 &35.0±0.0 &47.1±0.1 \\
CFA ($K$=1) ($h$) &44.9±0.0 &41.2±0.1 &25.6±0.1 &54.4±0.1 &33.3±0.1 &30.6±0.0 &33.6±0.1 &44.0±0.0 \\
CFA ($K$=1) ($f$) &45.3±0.1 &41.7±0.2 &26.3±0.0 &60.6±0.2 &34.2±0.1 &31.2±0.0 &34.5±0.1 &45.3±0.0 \\
CFA ($K$=3) ($h$) &44.8±0.1 &41.2±0.1 &25.6±0.1 &54.4±0.2 &33.2±0.1 &30.5±0.0 &33.5±0.1 &43.9±0.0 \\
CFA ($K$=3) ($f$) &45.2±0.1 &42.2±0.1 &25.7±0.1 &55.6±0.1 &34.1±0.1 &31.2±0.1 &33.9±0.0 &44.6±0.0 \\
CFA ($K$=5) ($h$) &44.8±0.0 &41.1±0.0 &25.6±0.1 &54.3±0.2 &33.2±0.1 &30.6±0.0 &33.6±0.1 &43.9±0.0 \\
CFA ($K$=5) ($f$) &51.4±0.4 &50.6±0.7 &28.3±0.3 &69.2±1.3 &43.9±0.2 &38.9±0.2 &38.4±0.2 &52.3±0.1 \\
\bottomrule
\end{tabular}
\caption{Detailed experiment results of Table \ref{table_ablation_study} (Top-1 error rate). (h): Hidden representation is normalized before calculating statistics. (f): Hidden representationis not normalized.}
\label{table_ablation_study_detail}
\end{table*}

